\RequirePackage{fix-cm}
\documentclass[preprint,12pt]{elsarticle}
\biboptions{numbers,sort&compress}
\usepackage[
letterpaper, top=1in, bottom=1in, left=1in, right=1in]{geometry}
\usepackage{amsmath, amssymb, amsthm}
\usepackage{booktabs}
\usepackage{graphicx}
\usepackage{subcaption}

\makeatletter
\newcommand{\tablefontsize}{%
  \normalsize
  \fontsize{\dimexpr\f@size pt*9/10\relax}%
           {\dimexpr\baselineskip*9/10\relax}\selectfont}
\makeatother
\AddToHook{env/table/begin}{\tablefontsize}
\AddToHook{env/table*/begin}{\tablefontsize}
\DeclareCaptionFont{tablefontsize}{\tablefontsize}
\usepackage{algorithm}
\usepackage{algpseudocode}
\usepackage[colorlinks=true, linkcolor=blue, citecolor=blue]{hyperref}

\theoremstyle{definition}
\theoremstyle{remark}
\newcommand{\E}{\mathbb{E}}
\newcommand{\R}{\mathbb{R}}
\newcommand{\Var}{\mathrm{Var}}
\newcommand{\dd}{\,\mathrm{d}}

\newcommand{\maybefigp}[2]{%
  \IfFileExists{#1}{\includegraphics[width=#2]{#1}}{%
  \fbox{\parbox{#2}{\centering\vspace{2em}pending run:\\
  \texttt{\detokenize{#1}}\vspace{2em}}}}}
\newcommand{\exfourfig}[2]{\maybefigp{figs/ex4/#1}{#2}}
\newcommand{\exgletwofig}[2]{\maybefigp{figs/ex2gle/#1}{#2}}
\newcommand{\exsolfig}[2]{\maybefigp{figs/ex3sol/#1}{#2}}

\journal{Journal of Computational Physics}

\begin{document}

\begin{frontmatter}

\title{Memory-Conditioned Diffusion Model for Generalized
Langevin Dynamics\tnoteref{fn1}}

\tnotetext[fn1]{{\bf Notice}:
This manuscript has been authored by UT-Battelle, LLC, under contract DE-AC05-00OR22725 with the US Department of Energy (DOE). The US government retains and the publisher, by accepting the article for publication, acknowledges that the US government retains a nonexclusive, paid-up, irrevocable, worldwide license to publish or reproduce the published form of this manuscript, or allow others to do so, for US government purposes. DOE will provide public access to these results of federally sponsored research in accordance with the DOE Public Access Plan.}

\author[ornl]{Minglei Yang\corref{cor1}}
\ead{yangm@ornl.gov}

\author[utk]{Sicheng He}
\ead{sicheng@utk.edu}

\cortext[cor1]{Corresponding author}

\affiliation[ornl]{organization={Fusion Energy Division, Oak Ridge National Laboratory},
            city={Oak Ridge},
            state={TN},
            country={USA}}

\affiliation[utk]{organization={Department of Mechanical and Aerospace Engineering, University of Tennessee},
            city={Knoxville},
            state={TN},
            country={USA}}

\begin{abstract}
Generalized Langevin equations describe non-Markovian dynamics in which the evolution of resolved variables depends on their past.
We propose a memory-conditioned diffusion method for learning stochastic flow maps of these dynamics from observed trajectories, without identifying a memory kernel or reconstructing unresolved variables.
A compact, recursively updated bank of exponential filters enables the flow map to retain predictive history over multiple time scales without conditioning on long observation windows.
The next-step distribution is conditioned on the current observation and this memory state, whose storage and update costs are independent of the history length for a fixed bank size.
Predictive criteria guide the memory budget, with reference-assisted selection in the vector benchmark, and an optional linear projection further reduces the conditioning dimension.
A kernel-based score estimator generates conditional samples without training a score network, and these samples are used to train a neural flow map for autoregressive simulation.
Three numerical examples assess long-memory retention at small conditioning dimension, predictive compression in coupled vector dynamics, and non-Gaussian conditional distributions and intermittent events.
The non-Gaussian example reproduces conditional asymmetry and burst statistics in a stochastic model of the plasma scrape-off layer.
\end{abstract}

\begin{keyword}
generalized Langevin equation \sep non-Markovian stochastic dynamics \sep
training-free conditional diffusion \sep
stochastic flow map \sep fading memory \sep predictive state
\end{keyword}

\end{frontmatter}

\section{Introduction}
\label{sec:intro}

When a complex system is observed through only a few resolved variables, their evolution is generally not Markovian: unresolved degrees of freedom retain information about the past and feed it back into the future.
The Mori--Zwanzig formalism makes this mechanism explicit by expressing reduced dynamics in terms of an instantaneous contribution, a memory term, and an orthogonal fluctuating force \citep{zwanzig1961, mori1965, zwanzig2001}.
The generalized Langevin equation (GLE) is a canonical continuous-time realization and is widely used in molecular coarse graining and anomalous diffusion \citep{lei2016datadriven, mckinley2018anomalous}.
We focus on learning GLE dynamics from trajectories of the resolved variables.
The same formulation also applies to partially observed pulse processes, including models of intermittent scrape-off-layer fluctuations \citep{garcia2012stochastic, theodorsen2017filtered}.

When a closed reduced equation and its latent variables are known, history dependence can often be handled by augmenting the state and applying standard simulation methods; exponential memory kernels, for example, admit extended-variable Markovian representations \citep{ceriotti2010colored, baczewski2013numerical}.
When only resolved trajectories are available, the latent state and the conditional distribution must instead be represented from data.
A useful learned model must therefore retain the predictive information in the observed history, represent the full conditional distribution rather than only its mean, and remain stable when propagated autoregressively.

One line of work identifies an explicit non-Markovian reduced model before simulation.
Jung et al.~\citep{jung2018generalized} construct and numerically integrate coarse-grained particle models with distance-dependent memory kernels reconstructed from fine-grained simulations, addressing systems with incomplete time-scale separation.
Memory kernels have been inferred from correlation functions and regularized inverse formulations \citep{lei2016datadriven, lang2026kernels}, including neural estimators designed for long-lived kernels \citep{winter2023deep}; likelihood-based formulations parameterize extended models with hidden auxiliary variables \citep{vroylandt2022likelihood}; and other approaches construct discrete non-Markovian closures, learn history-encoding features, or model state-dependent memory directly from resolved trajectories \citep{lin2021datadriven, she2023nonmarkovfeatures, ge2024statedependent}.
These methods provide an interpretable reduced equation, but they also require choosing a model structure for the memory and the fluctuating terms.

A second line learns the reduced evolution directly from trajectories.
Finite history windows and recurrent networks have been used to represent memory in reduced and partially observed dynamics \citep{ma2019memory, fu2020memory, harlim2021missing}, including long short-term memory (LSTM) networks \citep{hochreiter1997lstm}.
Separately, stochastic flow-map methods use generative sub-maps to represent unresolved randomness \citep{chen2024flowmap, chen2024multiscale}.
More recently, flow matching has been used to model non-Markovian and non-Gaussian flux distributions in coarse-grained stochastic particle systems \citep{siddani2026flowmatching}.
These approaches motivate two separate questions: how to retain predictive history in a compact state, and how to represent the conditional distribution given that state.
A recurrent representation addresses the first question, but the distributions it can generate also depend on its output family.

Training-free conditional diffusion provides a complementary route for learning stochastic flow maps \citep{liu2025trainingfree}.
For a Markovian SDE, the score of the one-step conditional distribution has an explicit representation that can be estimated from neighboring trajectory pairs without training a score network.
Solving the associated reverse probability-flow equation then generates labeled conditional samples, which are distilled into a supervised flow-map network.
The framework has since been extended to bounded domains with particle escape and to parameter-dependent SDEs \citep{yang2026bounded, yang2026parameter}, and dimension-dependent reverse-ODE discretization estimates have been developed for Gaussian-mixture targets \citep{wang2026error}.
These constructions condition on the instantaneous state, together with any prescribed parameters, and therefore pool observations that share the same current state even when their histories imply different next-step laws.

Our central contribution is a compact, recursively updated memory representation that enables conditional generative flow maps to preserve long-time dependence without conditioning on long observation windows.
We realize this construction by extending training-free conditional diffusion to condition on a multiscale bank of exponential filters of the observed increments.
The bank retains history over prescribed time scales, with storage and update costs independent of the history length for a fixed bank size.
One-step prediction and closed-loop correlation criteria guide the choice of filter rates and bank size, with reference-assisted choices identified in the vector benchmark.
A linear predictive reduction can further compress the conditioning variable before score estimation.
The memory representation, conditional sampler, and trained flow map are evaluated separately to distinguish representation, sampling, and deployment errors.
The construction combines established ideas in a conditional sampling framework.
Exponential filters are related to classical smoothing and extended-variable memory representations \citep{ceriotti2010colored, baczewski2013numerical}.
The predictive reduction uses linear covariance structure, as in reduced-rank regression \citep{izenman1975reduced}.
Here the purpose is to obtain a low-dimensional conditioning variable for a nonparametric score estimator while retaining a recursive update of the full memory bank.
The numerical examples assess three complementary aspects of the construction.
The scalar multiscale GLE tests long-memory retention at small conditioning dimension; the coupled vector GLE tests predictive compression and memory-induced coupling; and the stochastic scrape-off-layer model tests non-Gaussian conditional distributions and intermittent events.
Comparisons with memoryless, raw-history, and recurrent models help assess the roles of history representation and conditional output family.

Section~\ref{sec:problem} formulates the non-Markovian flow-map learning problem, and Section~\ref{sec:method} presents the memory model, conditional sampler, relevance metric, and distillation procedure.
Section~\ref{sec:examples} presents the numerical examples, and Section~\ref{sec:conclusions} concludes.
The appendices provide details of the memory representation, reverse-ODE discretization, and conditional score construction.

\section{Problem setting}
\label{sec:problem}

Our aim is to learn stochastic dynamics from trajectories of observed variables when unresolved degrees of freedom make their evolution depend on the past.
Generalized Langevin equations describe this history dependence through a memory force and temporally correlated fluctuations.
We consider the following system for the position $x(t)\in\R^d$ and velocity $v(t)\in\R^d$ \citep{didier2022harmonic}:
\begin{equation}
\label{eq:gle}
\begin{aligned}
\dd x(t) &= v(t)\dd t,\\
M\dd v(t) &= \left[F(x(t))-\Gamma_0v(t)
-\int_0^t K(t-s)v(s)\dd s+\eta(t)\right]\dd t
+\Sigma_0\dd W_t.
\end{aligned}
\end{equation}
Here $M$ is a positive-definite mass matrix and $F$ is a deterministic force.
The term $-\Gamma_0v(t)$ represents instantaneous drag, while the convolution with the matrix-valued kernel $K$ describes the influence of past velocities.
Random forcing consists of a colored force $\eta(t)$ --- which this GLE formulation models as a centered stationary Gaussian process, an assumption specific to Eq.~\eqref{eq:gle} and not shared by the Poisson-pulse process of Section~\ref{sec:ex3sol} --- and a direct white-noise contribution with amplitude matrix $\Sigma_0$, where $W$ is a standard Brownian motion.
For a stationary formulation, the memory integral extends over the entire past, with lower limit $-\infty$.

Related generalized Langevin models, including extensions with configuration-dependent memory outside the fixed convolution form of Eq.~\eqref{eq:gle}, are used in molecular coarse-graining and memory-kernel reconstruction \citep{li2015incorporation, jung2018generalized}, and measured hydrodynamic memory of this kind produces resonances in Brownian motion \citep{franosch2011resonances}.
In viscoelastic microrheology, such models connect the motion of trapped Brownian particles to the mechanical properties of the surrounding fluid \citep{paul2021bayesian}.
Their ergodic and long-time behavior has also been studied mathematically \citep{ottobre2011asymptotic, mckinley2018anomalous}.
Equation~\eqref{eq:gle} restricts attention to linear convolution memory; more general reduced equations can have nonlinear or state-dependent memory \citep{zwanzig2001}.
For an equilibrium bath with symmetric $K(t)$ and symmetric nonnegative $\Gamma_0$, the fluctuation--dissipation relations \citep{kubo1966fluctuation} are
\[
\Sigma_0\Sigma_0^\top=2k_BT\Gamma_0,
\qquad
\E[\eta(t)\eta(s)^\top]=k_BT K(|t-s|),
\]
where $k_B$ is Boltzmann's constant and $T$ is the temperature.
Here the colored force is centered and Gaussian, independent of $W$, and initialized compatibly with equilibrium.
These equilibrium relations are not required by the proposed learning method.

We seek to generate trajectories that reproduce the conditional evolution of the observed variables, without identifying $K$ or reconstructing the unresolved bath state.
We therefore formulate the learning task in terms of the available observations and their history-conditioned next-step distribution.

Let $X$ denote the $d$-dimensional observed process, sampled at a prescribed interval $\Delta t$.
The symbol $X$ denotes the observable used for learning and is distinct from the physical position $x(t)$ in Eq.~\eqref{eq:gle}; for velocity observations, $X(t)=v(t)$.
For a continuous-time process, $X_n := X(t_n)$ denotes the state on the uniform mesh $t_n = n \Delta t$; for an intrinsically discrete process, $X_n$ denotes its $n$-th state.
The available data consist of $N_{\mathrm{traj}} \ge 1$ records sampled on the same mesh,
\begin{equation}
\label{eq:data}
\mathcal{D}_{\mathrm{obs}} :=
\bigl\{\, X_0^{(i)}, X_1^{(i)}, \dots, X_L^{(i)}
\;\big|\; i = 1, \dots, N_{\mathrm{traj}} \,\bigr\},
\end{equation}
where $X_n^{(i)}$ denotes the state of the $i$-th trajectory at time $t_n$.
Only resolved observations enter the conditional estimator and flow-map training.
Known models can provide independent reference statistics for validation; any use of these references to select the memory representation is stated explicitly.

Because the current observation need not determine the next-step distribution, the learning target conditions on the observed history.
Writing $\mathcal{H}_n := (\dots, X_{n-1}, X_n)$ for the history of the resolved state up to time $t_n$, the resolved dynamics is governed by the history-conditioned one-step law
\begin{equation}
\label{eq:nonmarkov}
X_{n+1} \;\sim\; p\bigl(\,\cdot \mid \mathcal{H}_n\bigr),
\end{equation}
where $p(\,\cdot\mid\mathcal H_n)$ denotes a regular conditional distribution given $\sigma(X_k : k \le n)$, which exists on the Polish state spaces considered here, and the dependence on the fixed step $\Delta t$ is left implicit.
No parametric form is assumed for this conditional law.
Each data record supplies a finite portion of the history $\mathcal H_n$.

A model conditioned only on $X_n$ estimates the Markovian projection $p(\,\cdot\mid X_n)$, which averages over histories with different predictive laws.
We instead seek a generative flow map conditioned on a recursively computable summary of $\mathcal H_n$, to represent the full conditional distribution and generate new trajectories by repeated sampling and memory updates.

We make two assumptions to support learning this law from finite trajectory records.
\begin{enumerate}
\renewcommand{\labelenumi}{(A\arabic{enumi})}
\renewcommand{\theenumi}{A\arabic{enumi}}
\item\label{as:stat} \emph{Stationarity and ergodicity}: the law in Eq.~\eqref{eq:nonmarkov} is invariant under a shift of the time index, the process is statistically stationary, and time averages along a trajectory converge to the corresponding ensemble averages.
For a spatially homogeneous process, whose state need not be stationary, this is required only of the translation-invariant observables, namely the increments and the memory features computed from them.
\item\label{as:fade} \emph{Fading memory}: the predictive influence of the remote past on the law of the next state diminishes with the time lag.
This is a predictive analogue of fading memory for input--output operators \citep{boyd1985fading}.
Generalized Langevin equations with stable Prony-series kernels are a familiar example: their memory terms weight the remote past by decaying exponentials, and the same structure admits a finite-dimensional auxiliary-variable representation \citep{ceriotti2010colored, baczewski2013numerical}.
\end{enumerate}
Stationarity and ergodicity allow trajectory tuples to estimate a common conditional law.
Fading memory motivates a finite summary of the observed history, but does not by itself guarantee that such a summary accurately determines the next-step distribution.

\section{Memory-conditioned diffusion and flow-map learning}
\label{sec:method}

We build on the training-free conditional diffusion framework for stochastic flow-map learning \citep{liu2025trainingfree} and its extensions \citep{yang2026bounded, yang2026parameter}.
That framework estimates a conditional score from trajectory data, generates increment samples by reverse diffusion, and trains a neural network to reproduce the sampler.
Here we extend the conditioning to include a recursively updated summary of the observed history.
The central construction combines multiscale memory filters, predictive selection and optional compression of their outputs, and conditional sampling within an autoregressive simulator.
We first describe the memory construction, then summarize the inherited sampling and training steps with the modifications needed for history conditioning.

We write $X_n$ for the sampled observed process and $x_n$ for an observed or generated realization of $X_n$; lowercase $x_n$ in the learning formulation does not denote the physical position $x(t)$ in Eq.~\eqref{eq:gle}.
The prediction target is the observed-state increment $\Delta X_{n+1}:=X_{n+1}-X_n$, from which the next state is obtained by addition to $X_n$.
Throughout, \emph{training-free} refers specifically to score estimation and conditional sample generation: no score network is trained.

\subsection{A recursive memory model}
\label{sec:features}
Conditioning on the full history is impractical, and even a long finite window can make conditional estimation difficult because of its dimension.
We therefore compress it into a \emph{memory model}: a low-dimensional summary of the history, updated recursively along the trajectory,
\begin{equation}
\label{eq:memmodel}
m_n \;=\; \varphi(m_{n-1}, X_{n-1}, X_n).
\end{equation}
The summary is chosen to approximate the next-step law by
\begin{equation}
\label{eq:memapprox}
p\bigl(X_{n+1} \mid \mathcal{H}_n\bigr)
\;\approx\;
p\bigl(X_{n+1} \mid X_n, m_n\bigr).
\end{equation}

We realize $\varphi$ as a bank of exponential moving averages of the observed increments, following the classical exponential-smoothing recursion,
\begin{equation}
\label{eq:ema}
m_n^{(i)} \;=\; \rho_i\, m_{n-1}^{(i)} + \bigl(X_n - X_{n-1}\bigr),
\qquad i = 1, \dots, k,
\qquad\text{with}\quad \rho_i := e^{-\lambda_i \Delta t},
\end{equation}
with $m_0 = 0$ and rates $0 < \lambda_1 < \dots < \lambda_k$; the summary is $m_n := (m_n^{(1)}, \dots, m_n^{(k)}) \in \R^{kd}$.
Each filter accumulates past increments with exponential decay: small $\lambda_i$ retain longer histories, while large $\lambda_i$ emphasize recent changes.
The $k$ rates are logarithmically spaced over a selected interval $[\lambda_{\min},\lambda_{\max}]$.
They provide a set of history features and are not estimates of the physical memory-kernel parameters.
Compared with an LSTM, whose retention of history is learned through recurrent gates, the bank provides explicit memory scales without training recurrent weights.
This is useful when predictive information persists over widely separated time scales, since slow filters retain long histories by construction; the tradeoff is a fixed linear summary rather than a learned nonlinear memory representation.
The bank has $kd$ components and costs $O(kd)$ operations per update, independently of the trajectory length.
Training discards an initial transient, and simulation initializes the bank from an observed prehistory.

The rate interval and bank size are selected from the time scales the data resolve.
Writing $\ell_{\mathrm{sig}}$ for the largest lag at which the normalized autocorrelation of a declared diagnostic observable exceeds a sampling-noise threshold, candidate lower endpoints are log-spaced between $1/(\ell_{\mathrm{sig}}\Delta t)$ and $1/\Delta t$, and candidate upper endpoints are $\lambda_{\max}=c/\Delta t$ for a few prescribed constants $c$ of order one, since rates faster than the sampling rate act only as filters of the most recent increment.
Stage~A retains, for each candidate $k$, the bands whose held-out one-step prediction risk of a linear next-increment predictor given $(x_n,m_n)$ lies within a declared tolerance of the best; this risk is nearly flat across bands whenever slow memory contributes little to the next step.
Stage~B therefore rolls each retained linear surrogate forward autoregressively with its fitted residual covariance and selects the smallest $k$, and at that $k$ the band, whose closed-loop autocovariance error meets a preset tolerance; if none does, the smallest-error pair is retained and the failure is reported.
\ref{app:selection} gives the numerical selection settings; reference-based choices are identified with the examples.

Correlated filters can make the conditioning dimension larger than necessary.
An optional linear projection therefore retains the memory directions most predictive of future observations under a linear covariance diagnostic: after removing the part of the bank explained linearly by the current state, the singular spectrum of the whitened memory--future covariance defines a \emph{predictive rank} --- the number of memory directions resolved by this diagnostic --- and its leading directions give $m_n^{\mathrm{pred}}\in\R^{d_{\mathrm{pred}}}$.
The conditional model then receives $(x_n,m_n^{\mathrm{pred}})$, while the full bank continues to be updated by Eq.~\eqref{eq:ema}; the construction and its validation are given in \ref{app:prank}.
We assess the adequacy of the memory summary using prediction errors (\ref{app:suff}) and generated-trajectory statistics.

\subsection{Conditional sampling by diffusion}
\label{sec:tfdiff}

We apply the conditional sampler of \citep{liu2025trainingfree} to the current observation together with its memory summary:
\begin{equation}
\label{eq:coupling}
c_n \;=\; (x_n, m_n).
\end{equation}
When the predictive reduction of Section~\ref{sec:features} is applied, $m_n^{\mathrm{pred}}$ replaces $m_n$ in Eq.~\eqref{eq:coupling}.
Each condition $c_n$ is paired with the scaled increment $z_n = (x_{n+1} - x_n)\,\kappa_s$, where $\kappa_s > 0$ is fixed so that the increments have order-one variance.
Pooling over the trajectories of Eq.~\eqref{eq:data} gives $\{(c_i, z_i)\}_{i=1}^N$, whose conditional law we write $p(z \mid c)$; a generated $z$ gives the next state as $x + z/\kappa_s$.
This conditioning allows observations with similar current values but different predictive histories to contribute differently to the estimated next-step law.

For a fixed condition $c$, let $Z_0^c$ have the target density $p(\,\cdot\mid c)$.
Diffusion sampling introduces an artificial time $\tau$, distinct from the physical time $t$, and gradually adds Gaussian noise to the increment:
\begin{equation}
\label{eq:forward}
Z_\tau^c \mid Z_0^c = z_0 \;\sim\;
\mathcal{N}\bigl(\alpha_\tau z_0,\; \beta_\tau^2 I\bigr).
\end{equation}
Here $\alpha_\tau=1-\tau$ and $\beta_\tau^2=\tau$ for $0\leq\tau<1$, so the density $p_\tau(\,\cdot\mid c)$ approaches the standard normal as $\tau\uparrow1$.
We denote its conditional score by
\begin{equation}
\label{eq:scoredef}
S(z,\tau \mid c) \;:=\; \nabla_z \log p_\tau(z \mid c).
\end{equation}
To generate an increment, we reverse this transformation using the probability-flow ordinary differential equation (ODE) \citep{song2021score}:
\begin{equation}
\label{eq:reverseode}
\frac{\dd Z_\tau^c}{\dd \tau} \;=\;
b(\tau)\, Z_\tau^c - \frac{1}{2}\, \sigma^2(\tau)\,
S(Z_\tau^c,\tau \mid c),
\end{equation}
where $b(\tau)=\frac{d}{d\tau}\log\alpha_\tau$ and $\sigma^2(\tau)=\frac{d}{d\tau}\beta_\tau^2-2b(\tau)\beta_\tau^2$.
With the exact score, the ideal reverse flow maps standard-normal draws to the target distribution in the endpoint limit.
Numerical sampling uses a regularized finite-step approximation; the scheme and endpoint error are described in \ref{app:sampler}.

We estimate the score directly from the observed pairs $(c_j,z_j)$, without fitting a score network.
At a query condition $c^*$, we select its $J$ nearest neighbors and weight each increment by both the similarity of its condition to $c^*$ and its compatibility with the current diffused state $z$.
This gives
\begin{equation}
\label{eq:score}
\widehat S^{\mathrm{MC}}(z,\tau \mid c^*) \;:=\;
\sum_{j \in \mathcal{J}(c^*)} w_j(z, \tau)\,
\frac{\alpha_\tau z_j - z}{\beta_\tau^2},
\end{equation}
where the weights combine the two kernels,
\begin{equation*}
w_j(z, \tau) \propto
\underbrace{e^{-\frac{\|z - \alpha_\tau z_j\|^2}{2\beta_\tau^2}}}_{\text{diffusion kernel}}
\;
\underbrace{e^{-\frac{d(c^*, c_j)^2}{2\nu^2}}}_{\text{conditioning kernel}},
\end{equation*}
where $\mathcal{J}(c^*)$ is the index set of the $J$ nearest neighbors of $c^*$ under the metric $d$ defined below, and the weights are normalized so that $\sum_{j \in \mathcal{J}(c^*)} w_j = 1$.
Solving Eq.~\eqref{eq:reverseode} with the estimator in Eq.~\eqref{eq:score} in place of the exact score transports latent draws $\xi \sim \mathcal{N}(0, I)$ to approximate samples of $p(\,\cdot \mid c^*)$.
The derivation of this estimator is given in \ref{app:score}.
The \emph{memoryless baseline} replaces the conditioning of Eq.~\eqref{eq:coupling} by its instantaneous counterpart, $c_n = x_n$, which recovers the Markovian method of \citep{liu2025trainingfree}.

The state and memory components of $c$ have different scales and can be strongly correlated, and concentration of distances in high dimensions can reduce the discrimination of nearest-neighbor queries \citep{beyer1999nearest}.
We therefore whiten the condition, $u=\Sigma_c^{-1/2}(c-\bar c)$, where $\bar c$ and $\Sigma_c$ are the empirical mean and covariance of the conditioning data with eigenvalues floored at a small positive value, and fit the centered increment by least squares, $\E[z\mid u]-\bar z\approx A_{\mathrm{reg}}^\top u$.
Writing $A_{\mathrm{reg}}=U_{\mathrm{reg}}\,\mathrm{diag}(s_{\mathrm{reg}})\,V_{\mathrm{reg}}^\top$ for its singular value decomposition, we set $\tilde U:=U_{\mathrm{reg}}\,\mathrm{diag}(s_{\mathrm{reg}})/s_{\mathrm{reg},\max}$ and use the distance
\begin{equation}
\label{eq:metric}
d(c,c')^2=\bigl\|\tilde U^\top(u-u')\bigr\|^2
+\varepsilon^2\|u-u'\|^2,
\end{equation}
with $\varepsilon>0$; if $s_{\mathrm{reg},\max}=0$, the supervised term is omitted.
The first term emphasizes directions that predict the conditional mean; the second keeps the distance sensitive to all conditioning directions.
Neighborhood quality is monitored through the radius of each query's neighbor set and the effective sample size of its normalized weights, and the bandwidth $\nu$ balances localization in the conditioning variable against the number of points contributing appreciably to the estimate.
Sampler settings are reported with each example.

\subsection{Neural flow-map training and simulation}
\label{sec:principles}

To avoid repeatedly solving the reverse ODE during long simulations, we use the supervised flow-map training procedure of \citep{liu2025trainingfree} with the history-dependent condition $c$.
The neural flow map $G_\theta:\R^{d_c}\times\R^d\to\R^d$, where $d_c$ is the conditioning dimension, reproduces the sampler's output from $c$ and a Gaussian input $\xi$.
The observed pairs $(c_i,z_i)$ do not specify corresponding Gaussian inputs; the diffusion sampler supplies these input--output training pairs.
We draw $Q$ condition--latent pairs, with $c_q$ sampled from the conditioning data and $\xi_q \sim \mathcal{N}(0,I)$ independently; a condition may be paired with more than one latent.
For each pair, Eq.~\eqref{eq:reverseode}, with $\widehat S^{\mathrm{MC}}$ in place of the exact score, produces a scaled-increment label $z_q^{\mathrm{TF}}$.

We use a fully connected network with hyperbolic-tangent activations and a linear output layer.
Its parameters $\theta$ are trained by least squares on the generated labels,
\begin{equation}
\label{eq:loss}
\min_\theta \; \frac{1}{Q} \sum_{q=1}^{Q}
\bigl\| D_z^{-1}\bigl(G_\theta(c_q, \xi_q) - z_q^{\mathrm{TF}}\bigr) \bigr\|^2 .
\end{equation}
Here $D_z$ is the diagonal matrix of label standard deviations, floored away from zero, accounting for the output standardization used in training.
Network inputs are also standardized, and $G_\theta$ denotes the complete map after undoing output standardization.
The Gaussian input $\xi$ allows the network to generate different increments at the same condition; the regression learns the sampler's output for each input pair, rather than only the conditional mean.
Network sizes and optimization settings are reported with the examples.

The trained network is deployed autoregressively.
Starting from a state and memory bank computed from an observed prehistory, we form the condition $c_n$ of Eq.~\eqref{eq:coupling}, draw $\xi_n \sim \mathcal{N}(0,I)$, and set
\begin{equation}
\label{eq:rollout}
x_{n+1} \;=\; x_n + G_\theta(c_n, \xi_n)/\kappa_s ,
\qquad
m_{n+1}^{(i)} \;=\; \rho_i\, m_n^{(i)} + (x_{n+1} - x_n),
\quad i = 1, \dots, k,
\end{equation}
so that the memory summary is updated from the generated increments.
The cost per step is one network evaluation plus the $O(kd)$ bank update, independently of the trajectory length.
The complete workflow is summarized in Algorithm~\ref{alg:workflow}.

\begin{algorithm}[H]
\caption{Memory-conditioned flow-map learning and deployment}
\label{alg:workflow}
\begin{algorithmic}[1]
\Require observed training trajectories $\mathcal{D}_{\mathrm{obs}}$ and an observed prehistory for rollout initialization
\Ensure trained flow map $G_\theta$ and a generated trajectory $\{x_n\}$
\Statex \textbf{Construct the conditioning data}
\State select the rate band and bank size using the predictive criteria
of Section~\ref{sec:features}, with any reference-based choices stated
\State compute the full bank $m_n$ along the training trajectories by
Eq.~\eqref{eq:ema} and discard the feature burn-in
\State form $c_n$ from the current observation and memory, applying
the selected reduction when used
\State pair $c_n$ with $z_n=\kappa_s(x_{n+1}-x_n)$
\State fit the conditioning metric $d$ in Eq.~\eqref{eq:metric}
\Statex \textbf{Generate labels and train the flow map}
\For{each sampled condition--latent pair $(c_q,\xi_q)$}
\State select the $J$ nearest training tuples to $c_q$ under $d$
\State integrate the regularized reverse probability-flow ODE with
Eq.~\eqref{eq:score} to obtain $z_q^{\mathrm{TF}}$
\EndFor
\State train $G_\theta$ on the resulting condition--latent--label
pairs by Eq.~\eqref{eq:loss}
\Statex \textbf{Generate a trajectory}
\State set $x_0$ to the last state of the initial prehistory and
compute $m_0$ from that prehistory using Eq.~\eqref{eq:ema}
\For{$n=0,1,\dots$ until the desired trajectory length}
\State form $c_n$ using the same conditioning variables as in training
and draw $\xi_n\sim\mathcal N(0,I)$
\State set $z_n^{\mathrm{gen}}=G_\theta(c_n,\xi_n)$ and
$x_{n+1}=x_n+z_n^{\mathrm{gen}}/\kappa_s$
\State update the full bank $m_{n+1}$ by Eq.~\eqref{eq:ema} using
$x_{n+1}-x_n$ and apply the same predictive read-out used in training
\EndFor
\State \Return $G_\theta$ and $\{x_n\}$
\end{algorithmic}
\end{algorithm}

\section{Numerical examples}
\label{sec:examples}

We present three complementary tests of memory-conditioned flow-map learning.
Example~1 evaluates long-memory retention at small conditioning dimension, Example~2 examines coupled vector dynamics and predictive compression, and Example~3 evaluates non-Gaussian conditional distributions and intermittent-event statistics.

Unless stated otherwise, all examples share the same numerical settings.
The conditional sampler uses $J = 1024$ neighbors, conditioning bandwidth $\nu = 0.2$, metric floor $\varepsilon = 0.3$, and $n_{\mathrm{ODE}} = 400$ reverse-ODE steps, and generates $2\times10^5$ training labels ($1.8\times10^5$ in Example~2).
The flow map has three hidden layers of $128$ hyperbolic-tangent units and is trained with Adam at learning rate $10^{-3}$, batch size $4096$, and at most $3000$ epochs with validation-based early stopping.
The LSTM baselines use $128$ hidden units in one recurrent layer and are trained by one-step maximum likelihood; their training-context lengths and deployment warm-ups are stated with each example.

\subsection{Linear GLE with multiscale memory}
\label{sec:ex1}

This example tests whether a small memory bank retains the history information needed for conditional forecasts and long-time correlations when the forcing contains widely separated time scales.
We consider a scalar velocity $X_t$ with unit mass and no external force, governed by
\begin{equation}
\label{eq:ex1model}
\begin{gathered}
\dd X_t \;=\; \left[-\gamma X_t - c_K \int_{-\infty}^{t} e^{-\lambda_K (t-s)}
X_s \dd s + \sum_{j=1}^{4} F_{j,t}\right]\dd t,\\
\dd F_{j,t} \;=\; -\tfrac{1}{T_j} F_{j,t} \dd t
+ \sigma_j \sqrt{2/T_j}\, \dd W_{j,t} ,
\end{gathered}
\end{equation}
where $\gamma = 0.5$, $c_K = 2$, and $\lambda_K = 1$.
The four Ornstein--Uhlenbeck forces have correlation times $T_j \in \{0.5,\ 4,\ 30,\ 200\}$ and stationary standard deviations $\sigma_j \in \{0.6,\ 0.5,\ 0.45,\ 0.45\}$.
In Eq.~\eqref{eq:gle}, this corresponds to $\Sigma_0=0$ and $\eta_t=\sum_j F_{j,t}$, prescribed independently of the friction kernel without imposing equilibrium fluctuation--dissipation relations.
Introducing $\zeta_t = \int_{-\infty}^{t} e^{-\lambda_K(t-s)}X_s\dd s$ gives the six-dimensional linear Markovian embedding $Y=(X,\zeta,F_1,\ldots,F_4)$.

We generate data using the exact discrete transition,
\begin{equation}
\label{eq:ex1data}
\begin{gathered}
Y_{n+1} = A_d\, Y_n + \eta_n, \qquad\text{with}\quad A_d := e^{A \Delta t},\\
\eta_n \sim \mathcal{N}\bigl(0,\; \Sigma - A_d \Sigma A_d^\top\bigr),
\qquad Y_0 \sim \mathcal{N}(0, \Sigma).
\end{gathered}
\end{equation}
Here $A$ and $\Sigma$ are the embedding drift matrix and stationary covariance.
Only $X$ is retained as observed data.
We use $\Delta t=0.2$, with $100$ training trajectories and $20$ test trajectories, each containing $2\times10^4$ steps.
The forcing times span $2.5$ to $1000$ sampling intervals.
The embedding supplies exact references for all conditioning choices.
Stages~A and~B of Section~\ref{sec:features}, with the state autocovariance as the diagnostic observable that fixes $\ell_{\mathrm{sig}}$, select $k = 6$ filters over the rate band $[\lambda_{\min}, \lambda_{\max}] = [0.038, 15.0]$, with the individual rates placed by Eq.~\eqref{eq:rates}.
The sampler and neural flow map use the full six-filter bank.

We compare the memory model with a memoryless model, a raw-history model conditioned on $c_n=(x_n,\ldots,x_{n-6})$, and a Gaussian-head LSTM, all trained on the same data.
The first three share the diffusion pipeline; the raw-history model and EMA bank also have equal conditioning dimensions.
The LSTM receives the current state and increment, predicts the next increment's mean and variance, and samples during rollout.
For the main comparison, it is trained by one-step maximum likelihood with a $1000$-step context and initialized from a $3000$-step observed history; the separate training-context study in Table~\ref{tab:ex1robust}(b) uses a common $5000$-step initialization history.
Its Gaussian output family matches the exact conditional laws.

\begin{figure}[htbp]
\centering
\exfourfig{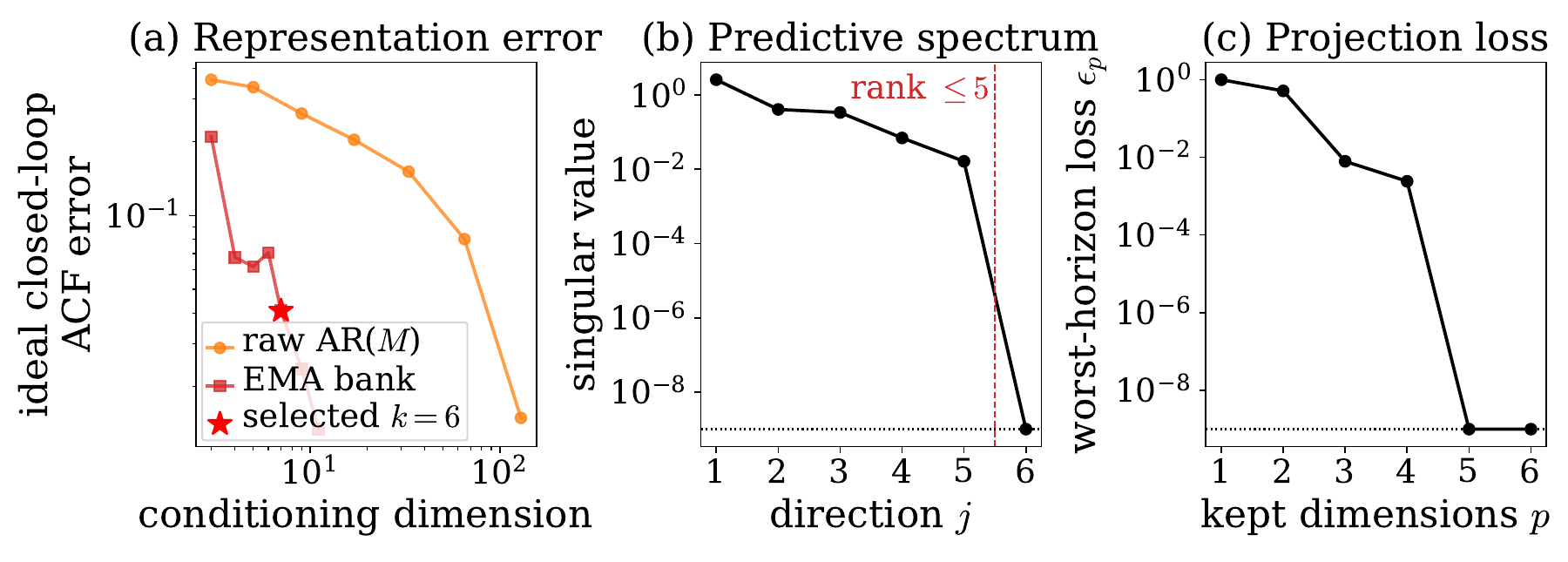}{0.92\textwidth}
\caption{Memory representation for Example~1, evaluated without learning error.
(a)~Closed-loop autocovariance error for a raw window and an EMA bank; the star marks the selected six-filter bank.
(b)~Predictive spectrum of the selected bank.
(c)~Worst-horizon loss after projection onto the first $d_{\mathrm{pred}}$ predictive coordinates.}
\label{fig:ex1mem}
\end{figure}

To assess the memory approximation separately, Figure~\ref{fig:ex1mem}(a) compares exact conditional models based on the selected bank and a raw window with $k=6$.
The selected six-filter bank has substantially smaller autocovariance error than short raw-history windows; comparable window accuracy requires a much larger conditioning dimension.
The predictive-coordinate diagnostic in panels (b)--(c) measures how much of this information lies in a smaller linear subspace.
The sharp drop in the predictive spectrum after the fifth direction agrees with the five hidden variables that influence $X$, and retaining these five directions introduces no measurable loss over the tested prediction horizons.

\begin{figure}[htbp]
\centering
\exfourfig{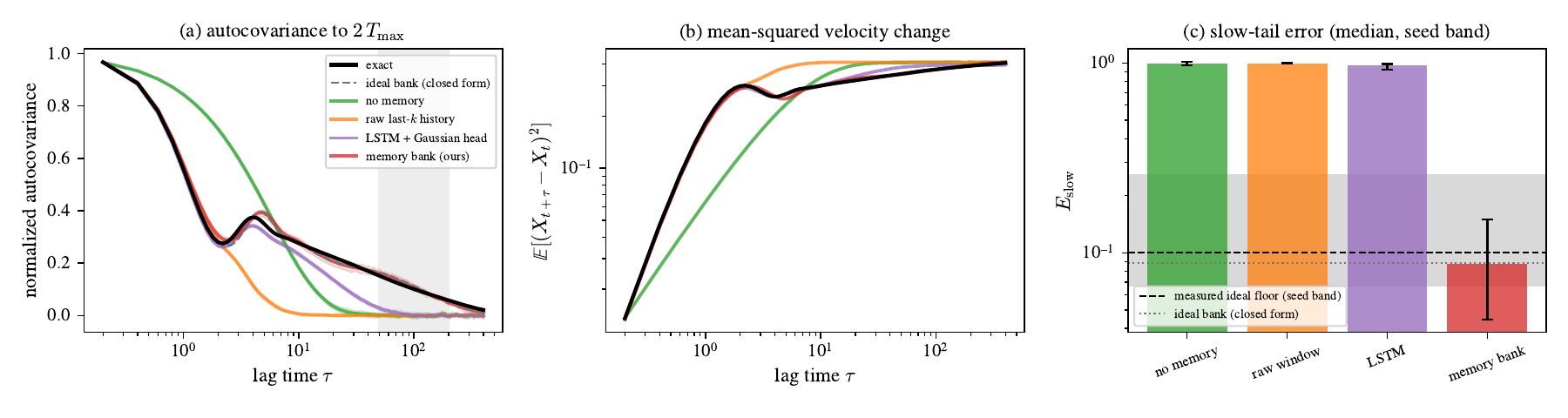}{0.9\textwidth}
\caption{Closed-loop statistics for Example~1.
(a)~Normalized autocovariance; the shaded interval defines the slow-tail error.
(b)~Mean-squared velocity change.
(c)~Slow-tail error, with the ideal EMA-bank error shown as a reference.}
\label{fig:ex1compare}
\end{figure}

We next evaluate the complete learned simulators through long autoregressive rollouts (Figure~\ref{fig:ex1compare}).
The memory model follows both the intermediate shoulder and the slowly decaying tail of the exact autocovariance.
The raw-history model loses the correlation after the first few time units, while the memoryless model decays on the wrong time scale.
The LSTM captures more of the intermediate behavior but still misses the slowest component.
The mean-squared velocity change $\E[(X_{t+\tau}-X_t)^2]$ shows the same differences; because the observed variable is a velocity, this statistic measures decorrelation of the velocity rather than a position displacement.
We quantify slow-memory accuracy using the exact autocovariance $C$ and rollout estimate $\widehat C$:
\begin{equation}
\label{eq:eslow}
E_{\mathrm{slow}} \;=\;
\frac{\sum_{\tau \in \mathcal{T}} \bigl| \widehat C(\tau) - C(\tau) \bigr|}
     {\sum_{\tau \in \mathcal{T}} \bigl| C(\tau) \bigr|},
\qquad \mathcal{T} = \{\, 50 \le \tau \le 200 \,\}.
\end{equation}
Here $\mathcal{T}$ contains the sampled lags between $50$ and $200$ in physical time units, rather than sampling steps.
A vanishing rollout autocovariance over this interval gives $E_{\mathrm{slow}}=1$.
The shoulder statistic is the maximum of $|\widehat C(\tau) - C(\tau)|/C(0)$ over physical time lags $0\le\tau\le30$.
Figure~\ref{fig:ex1compare}(c) shows that the memory model remains close to the ideal bank reference, while the raw-history model and LSTM have slow-tail errors near unity.

\begin{figure}[htbp]
\centering
\exfourfig{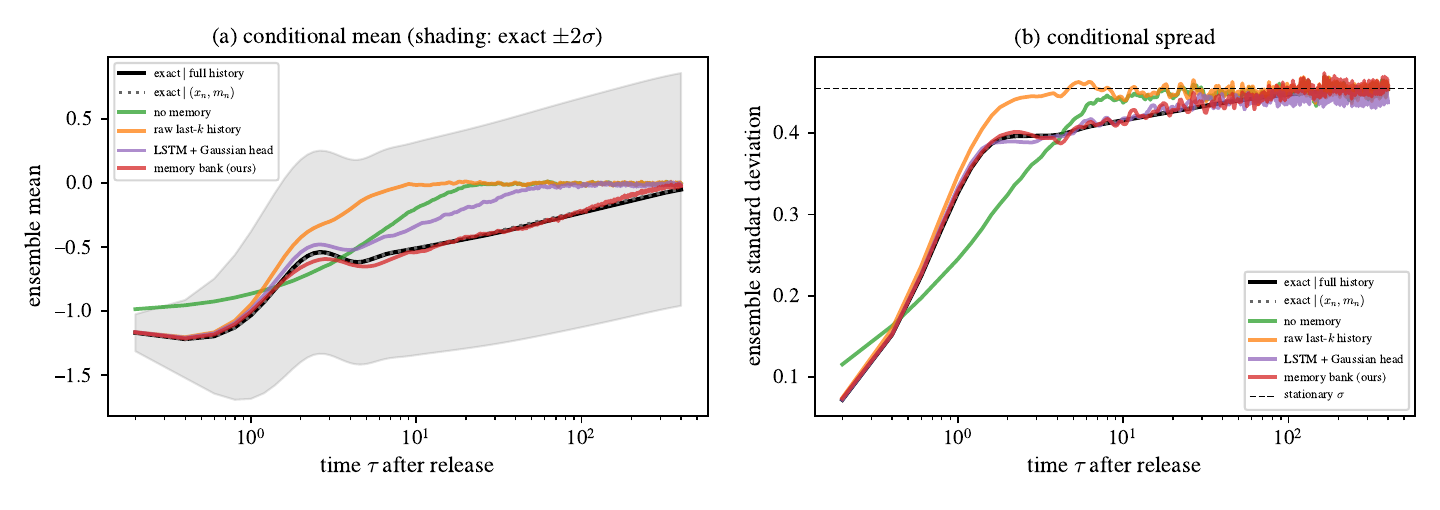}{0.9\textwidth}
\caption{Conditional forecasts for Example~1.
(a)~Conditional mean and (b)~conditional standard deviation of an ensemble initialized from the same observed history.
The solid black curve is the exact full-history forecast, and the dotted curve is the exact EMA-conditioning reference.}
\label{fig:ex1ens}
\end{figure}

Figure~\ref{fig:ex1ens} examines conditional forecasts initialized from the same observed history.
The chosen history has a large contribution from the slow hidden mode, making the effect of memory visible in the subsequent relaxation.
Panel (a) shows that the memory model follows the exact conditional mean through the slow rebound and remains close to the exact EMA-conditioning reference.
The raw-history and memoryless models relax too quickly, while the LSTM captures only part of the rebound.
Panel (b) shows that beyond the first several time units the four models predict a similar conditional spread; at shorter horizons the memoryless and window spreads deviate visibly from the exact one.
Over the remaining horizons, their main difference is therefore the history-dependent conditional mean, rather than the forecast variance.

To assess forecast accuracy beyond a single history, Table~\ref{tab:ex1robust}(a) compares conditional distributions over $32$ test histories using normalized squared Wasserstein distance.
The memory model has the smallest error among the learned models.
The table also tests whether the LSTM improves when its training context is increased from $1000$ to $2500$ and $5000$ steps.
The slow-tail error remains between $0.88$ and $1.00$ across the nine trained models, with no systematic improvement as the context grows.
Because the Gaussian output family is correctly specified here, these comparisons establish an empirical advantage for the tested configurations; they do not isolate whether the remaining gap stems from the recurrent representation, the optimization, or the one-step training objective.
As a sampler-refinement check --- doubling $n_{\mathrm{ODE}}$ both refines the reverse-ODE discretization and shrinks the endpoint regularization $h=1/n_{\mathrm{ODE}}$ (\ref{app:sampler}) --- repeating $16$ conditional queries with identical latent draws at $2n_{\mathrm{ODE}}$ steps gives, for the memory model, a median across queries of the mean squared paired-sample difference of approximately $1.52\times10^{-5}$, normalized by the exact conditional variance.
The selected memory bank preserves slow temporal dependence in both autoregressive simulation and conditional forecasting.

\begin{table}[H]
\centering
\caption{Robustness checks for Example~1.
(a)~Median normalized squared Wasserstein distance to the exact full-history conditional law at $\tau=30$, over $32$ test histories with $1000$ forecast samples per history.
The finite-sample floor compares a finite Gaussian sample with the analytic quantiles of the same Gaussian law; the exact EMA-conditioning error is evaluated analytically.
(b)~Slow-tail error for LSTMs trained with three context lengths; entries are the median and range over three training seeds.
All LSTMs in (b) use the same $5000$-step observed warm-up at deployment.}
\label{tab:ex1robust}
\begin{minipage}[t]{0.49\linewidth}
\vspace{0pt}
\centering
\setlength{\tabcolsep}{3pt}
\begin{tabular}{@{}lc@{}}
\toprule
\multicolumn{2}{c}{(a) Conditional forecast error at $\tau=30$} \\
\cmidrule(lr){1-2}
model & median $W_2^2/\sigma_{\mathrm{exact}}^2$ \\
\midrule
memoryless & $0.0405$ \\
raw-history model, $k=6$ & $0.0386$ \\
LSTM & $0.0145$ \\
memory model & $0.0040$ \\
exact EMA conditioning & $0.0004$ \\
finite-sample floor & $0.0027$ \\
\bottomrule
\end{tabular}
\end{minipage}\hfill
\begin{minipage}[t]{0.49\linewidth}
\vspace{0pt}
\centering
\setlength{\tabcolsep}{3pt}
\begin{tabular}{@{}lcc@{}}
\toprule
\multicolumn{3}{c}{(b) LSTM training-context study} \\
\cmidrule(lr){1-3}
model & \shortstack{training\\context} & $E_{\mathrm{slow}}$ \\
\midrule
LSTM & $1000$ & $0.978\ [0.913,\,0.979]$ \\
LSTM & $2500$ & $0.973\ [0.882,\,0.982]$ \\
LSTM & $5000$ & $0.991\ [0.980,\,0.992]$ \\
memory model & --- & $0.087$ \\
\bottomrule
\end{tabular}
\end{minipage}
\end{table}

\subsection{A two-dimensional viscoelastic GLE}
\label{sec:ex2gle}

GLEs with instantaneous drag and exponential memory describe Brownian motion in viscoelastic fluids \citep{paul2021bayesian}, and sums of exponential memory modes admit extended-variable formulations \citep{baczewski2013numerical}.
Here we use two coupled relaxation modes to test whether the method reproduces velocity correlations and conditional forecasts in a viscoelastic bath.
This benchmark tests vector dynamics with noncommuting memory modes and predictive compression using reference-assisted bank selection; the observation-based selection procedure is exercised in Examples~1 and~3.
We consider a particle in an equilibrium viscoelastic bath with unit mass, no external force, and $\Gamma_0=\gamma I$:

\begin{equation}
\label{eq:ex2model}
\begin{gathered}
\dd v_t=\left[-\gamma v_t-\int_{-\infty}^tK(t-s)v_s\dd s+\eta_t\right]\dd t
+\sqrt{2\gamma k_BT}\dd W_t,\\
K(t)=a_1e^{-t/\tau_1}u_1u_1^\top+a_2e^{-t/\tau_2}u_2u_2^\top,
\qquad \E[\eta_t\eta_s^\top]=k_BT K(|t-s|).
\end{gathered}
\end{equation}
Here $v_t\in\R^2$, $W_t$ is two-dimensional Brownian motion, and the centered Gaussian force $\eta$ is independent of $W$.
Both fluctuation--dissipation relations in Section~\ref{sec:problem} hold with $\Sigma_0=\sqrt{2\gamma k_BT}\,I$.
We set $u_1=(1,0)^\top$, $u_2=(1/2,\sqrt{3}/2)^\top$, $a_1=1$, $a_2=0.2$, $\tau_1=1$, $\tau_2=5$, $\gamma=1/9$, and $k_BT=1$.
The two rank-one kernel matrices do not commute, so no fixed rotation diagonalizes $K(t)$ at every lag.
The velocity autocovariance is
\begin{equation}
\label{eq:ex2vacf}
C(\tau)=\E[v(t+\tau)v(t)^\top].
\end{equation}
It satisfies $C(0)=I$ and has nonzero lagged cross-components with $C_{xy}(\tau)=C_{yx}(\tau)$.

A four-dimensional linear embedding supplies exact discrete transitions and transport statistics \citep{ceriotti2010colored, baczewski2013numerical}, and Kalman filtering gives conditional reference laws \citep{kalman1960filter}.
We retain only the sampled velocity $X_n=v_n:=v(t_n)$, using $32$ training and $8$ test trajectories of $2.5\times10^4$ steps each at $\Delta t=0.2$.
We use $k=5$ filters over $[0.0521,5]$, giving a $12$-dimensional state--memory vector.
Here the rate band is determined from the exact state autocovariance and the bank size from exact conditional and correlation references.
To reduce the conditioning dimension, we compress this vector to the increment predictor
\[
\widehat q_n=\widehat b+\widehat B_vv_n+\widehat B_mm_n\in\R^2
\]
fitted by ordinary least squares on the training tuples.
The exact bank-conditioned law is Gaussian with constant covariance, so its mean determines the increment distribution.
We set $c_n=\widehat q_n$ for both sampling and flow-map training and update the full bank during rollout.

\begin{figure}[htbp]
\centering
\exgletwofig{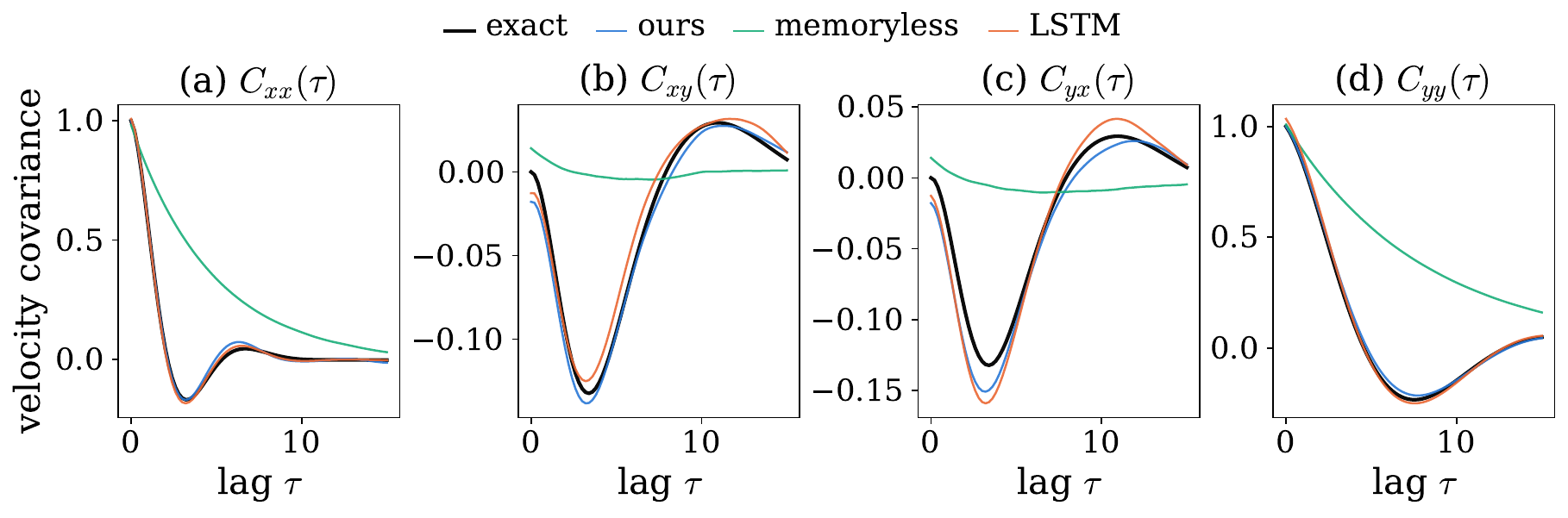}{0.98\textwidth}
\caption{Velocity-covariance matrix for Example~2 from long autoregressive rollouts.
Panels (a)--(d) show $C_{xx}$, $C_{xy}$, $C_{yx}$, and $C_{yy}$ against the exact curves; the nonzero lagged cross-covariance reflects the memory-induced coupling between the components (at equilibrium $C_{xy} = C_{yx}$).}
\label{fig:ex2vacf}
\end{figure}

We compare against memoryless diffusion and an LSTM with a full-covariance Gaussian output head.
The LSTM receives the current velocity and increment, uses a $500$-step training context, and is initialized from a $1000$-step observed history.
Its Gaussian output family matches this example's conditional laws.
We examine the entries of the covariance matrix in Eq.~\eqref{eq:ex2vacf}: diagonal entries measure temporal correlations within each velocity component, and off-diagonal entries measure correlations between components.
Figure~\ref{fig:ex2vacf} compares velocity covariances from $64$ autoregressive trajectories of $10^4$ steps per model.
The memory model and LSTM reproduce the negative correlation lobes and subsequent relaxation in all four entries.
The memoryless model instead produces slowly decaying diagonal correlations and nearly vanishing cross-correlations.

\begin{figure}[htbp]
\centering
\exgletwofig{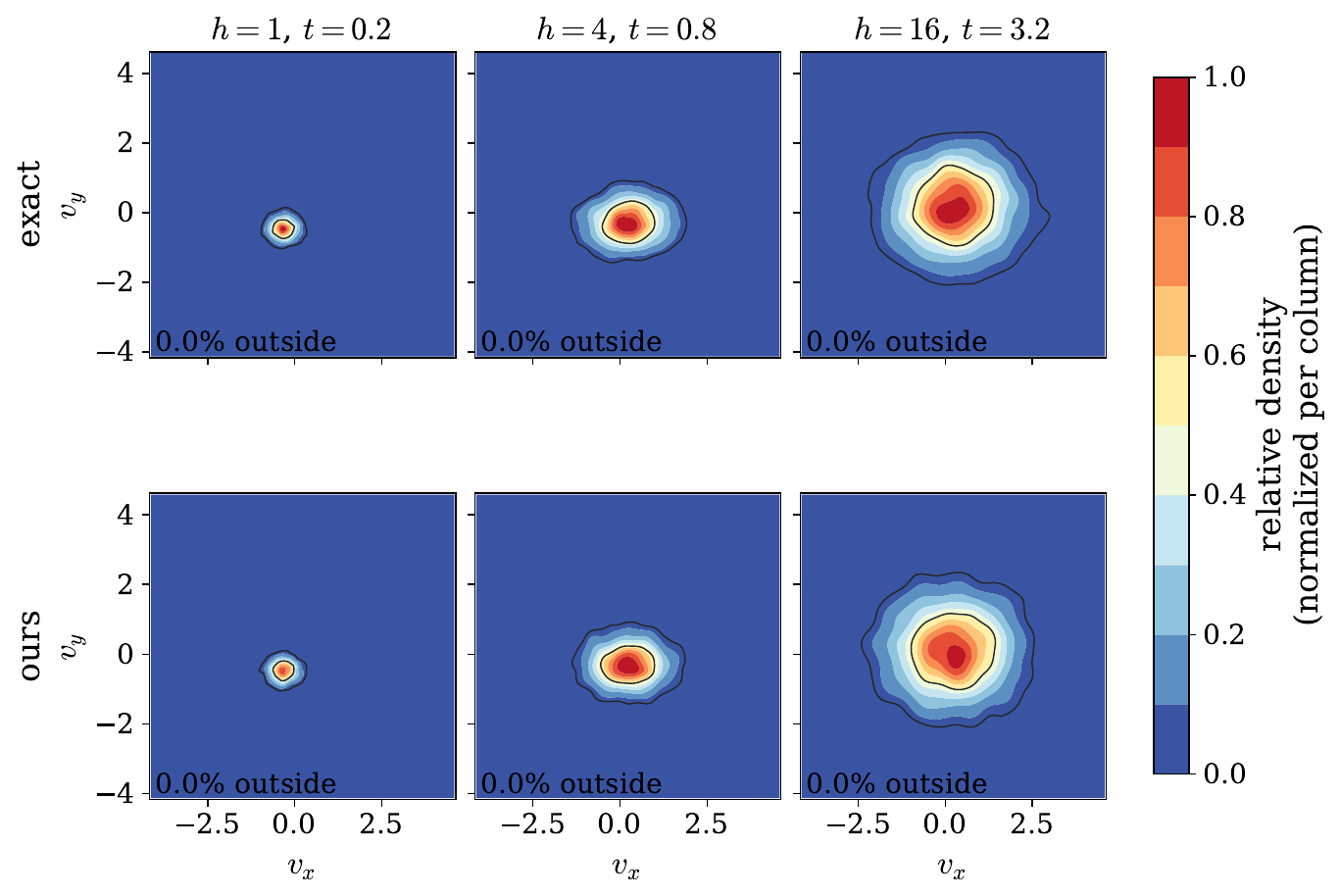}{0.92\textwidth}\\[2pt]
\exgletwofig{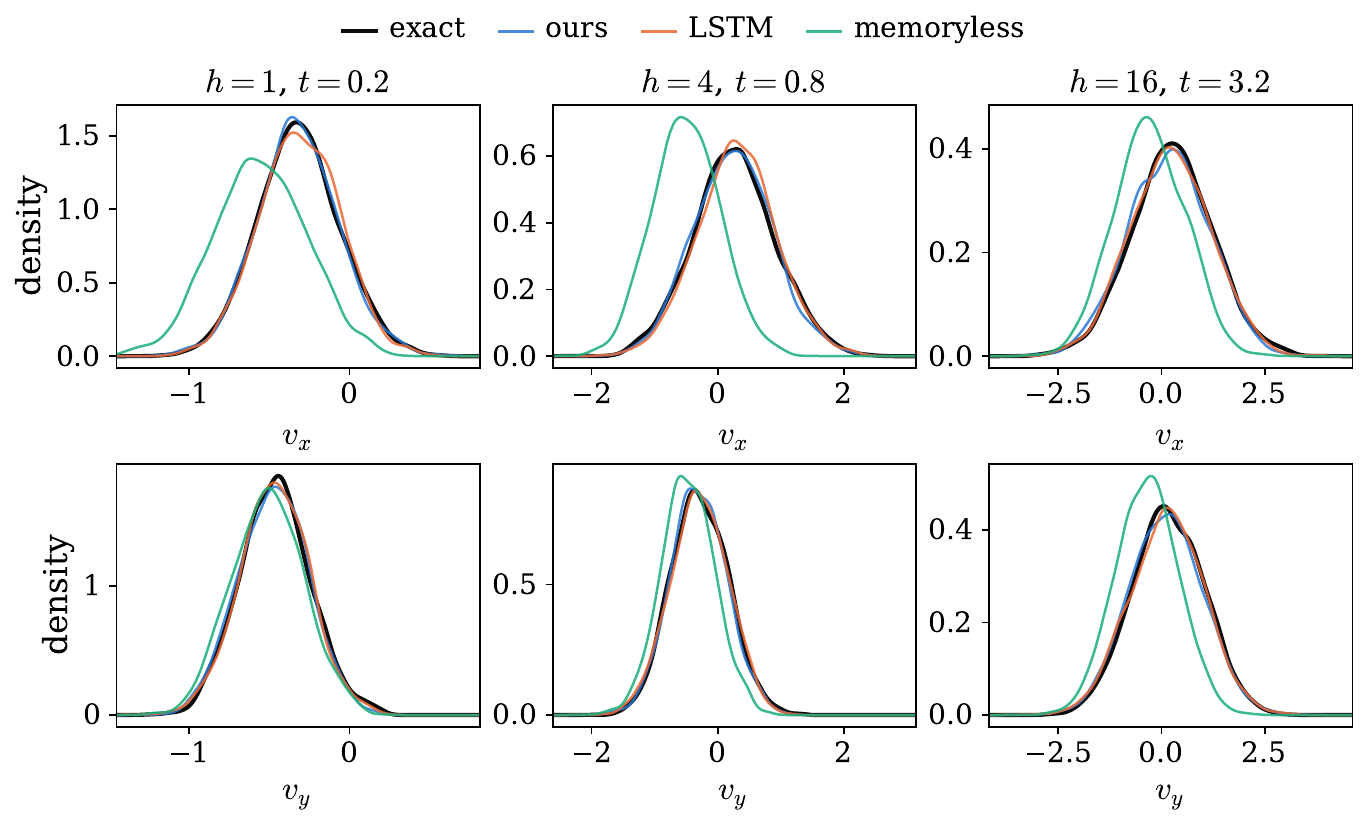}{0.92\textwidth}
\caption{Conditional forecast for Example~2 from one observed test history.
Top: joint velocity densities at $h=1,4,16$ for the exact conditional law (upper row) and memory model (lower row); contours enclose $50\%$ and $95\%$ of the smoothed in-frame mass.
Bottom: velocity marginals for all models.
Color scales and marginal axis limits are shared within each column.}
\label{fig:ex2evo}
\end{figure}

We next compare conditional forecasts with $3000$ paths from one test history, selected by reference predictive means among nearby current velocities, without using model errors.
The reference ensemble starts from the Kalman posterior of the embedding.
Figure~\ref{fig:ex2evo} shows two views of the conditional velocity distribution at forecast horizons of $h=1,4,16$ steps ($\Delta t=0.2$), with columns corresponding to successive horizons.
The upper panels show the joint density in the $(v_x,v_y)$ plane, with the reference ensemble in the first row and the memory model in the second; these panels display the location, spread, and dependence between the velocity components.
The memory model reproduces the reference density's location and spread as the ensemble evolves.
The lower panels show the marginal densities of $v_x$ and $v_y$ in the first and second rows, respectively, comparing the reference with the memory model, LSTM, and memoryless diffusion.
These curves assess each velocity component separately: both the memory model and LSTM match the marginal densities closely, while memoryless predictions are shifted, particularly in $v_x$.

\begin{figure}[htbp]
\centering
\exgletwofig{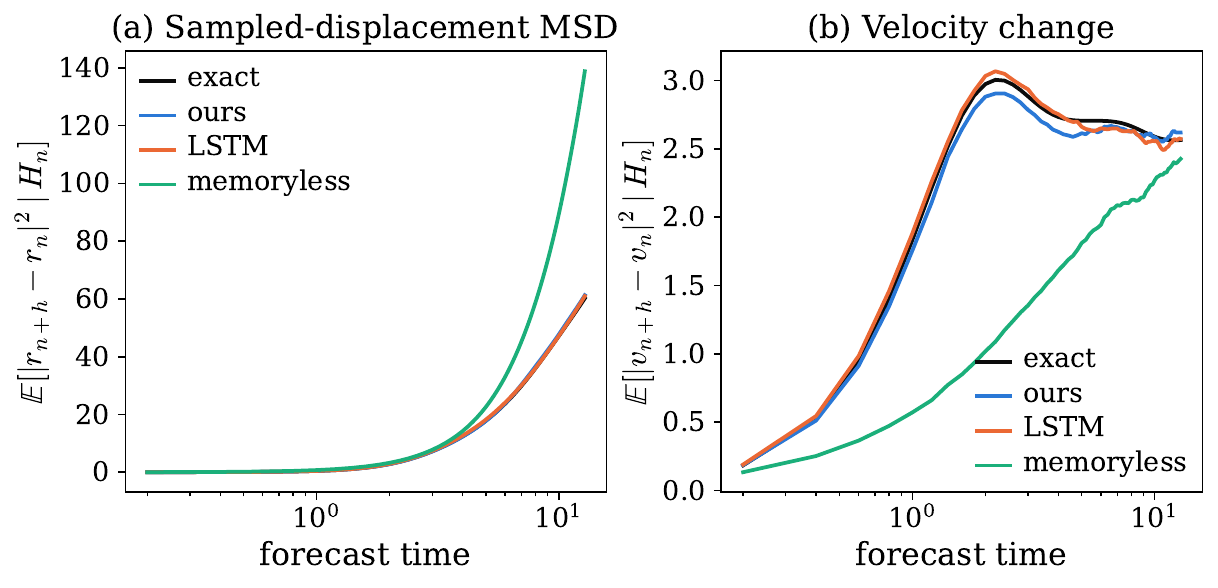}{0.85\textwidth}
\caption{Conditional second moments of the forecast of Figure~\ref{fig:ex2evo}.
(a)~Sampled-displacement mean-squared displacement and (b)~mean-squared velocity change, against the analytic conditional references (black).}
\label{fig:ex2msd}
\end{figure}

For the same forecast, Figure~\ref{fig:ex2msd} compares $\E[\|r_{n+h}-r_n\|^2\mid\mathcal H_n]$ and $\E[\|v_{n+h}-v_n\|^2\mid\mathcal H_n]$, with sampled displacement $r_{n+h}-r_n=\Delta t\sum_{j=1}^h v_{n+j}$.
The memory model and LSTM reproduce both the displacement growth and nonmonotone velocity-change relaxation.
The memoryless model overestimates displacement growth and misses the early maximum in the velocity-change statistic.

\subsection{Intermittent scrape-off-layer fluctuations}
\label{sec:ex3sol}

We apply the method to a stochastic model of intermittent scrape-off-layer (SOL) fluctuations \citep{garcia2012stochastic}, a physical application with memory and non-Gaussian bursts.
This example tests conditional prediction and the reproduction of burst statistics.
We use a filtered Poisson process with two populations of randomly arriving pulses --- our two-population, finite-rise extension of the one-population model analyzed in \citep{theodorsen2017filtered} --- observed as
\begin{equation}
\label{eq:ex3model}
\begin{aligned}
y_n &= \sum_{j=1}^{2} \sum_k A_{jk}\,\psi_j(t_n-t_{jk})
+\sigma_{\mathrm{obs}}\epsilon_n,\qquad \epsilon_n\overset{\mathrm{iid}}{\sim}\mathcal N(0,1),\\
\psi_j(s)&=\frac{\tau_{dj}}{\tau_{dj}-\tau_r}
\Bigl(e^{-s/\tau_{dj}}-e^{-s/\tau_r}\Bigr),\qquad s\geq0.
\end{aligned}
\end{equation}
Here $\psi_j(s)=0$ for $s<0$, $t_{jk}$ are Poisson arrival times, and the amplitudes $A_{jk}$ are exponentially distributed; the two arrival processes, all amplitudes, and the observation noise are mutually independent.
We set the decay times to $\tau_{d1}=1$ and $\tau_{d2}=10$, the common rise time to $\tau_r=0.1\tau_{d1}$, and the intermittency parameters $\gamma_j=\tau_{dj}/\tau_{wj}$ to $\gamma_1=2$ and $\gamma_2=1$, where $\tau_{wj}$ is the mean waiting time.
The amplitude means are $1$ and $\sqrt{101/55}$ for the two populations, fixed so that the populations contribute equal variance.

Only the combined signal $X_n=y_n$ is observed, at $\Delta t=0.5\tau_{d1}$, with observation-noise variance equal to $1\%$ of $\Var(y_n)$.
We use $400$ training and $120$ test records of $4000$ steps each.
The next-step distribution reflects both the decay of existing pulses and new arrivals.
An $8000$-particle filter on the hidden pulse state approximates the conditional reference law $p(y_{n+1}\mid y_{0:n})$.

Memory selection gives one filter with rate $0.26\,\tau_{d1}^{-1}$ and condition $c_n=(y_n,m_n^{\mathrm{pred}})\in\R^2$.
The baseline is a Gaussian-head LSTM trained by one-step maximum likelihood on the same records; a single training instance (seed $5$, fixed in advance) is used throughout.
It receives $(y_n,\Delta y_n)$, with $\Delta y_n=y_n-y_{n-1}$, and uses a $1000$-step training context and $1000$-step initialization history.

For one-step evaluation we group histories by the current observed signal amplitude $y_n$.
The \emph{quiet} group contains $64$ test histories with $y_n$ below its median, representing relatively low signal levels; the \emph{active} group contains $64$ histories with $y_n$ above its $80$th percentile, representing elevated signal levels associated with intermittent pulses.
These labels describe amplitude groups within the same stationary SOL model, rather than distinct plasma operating regimes.
For each test history, let $\widehat P_{\mathrm{model}}$ be the empirical distribution of $1024$ next-step samples from the model being evaluated, and let $\widehat P_{\mathrm{PF}}^{(1)}$ and $\widehat P_{\mathrm{PF}}^{(2)}$ be the empirical distributions of two independently drawn ensembles of the same size from the particle-filter predictive law conditioned on that history.
We measure the normalized excess error by
\[
E=\frac{W_2^2\!\left(\widehat P_{\mathrm{model}},\widehat P_{\mathrm{PF}}^{(1)}\right)
-W_2^2\!\left(\widehat P_{\mathrm{PF}}^{(2)},\widehat P_{\mathrm{PF}}^{(1)}\right)}
{s_{\mathrm{PF}}^2},
\]
where $W_2$ is the Wasserstein distance of order two and $s_{\mathrm{PF}}^2$ is the sample variance of the first reference ensemble.
The first term measures the discrepancy between model and reference samples; the second estimates the discrepancy caused by finite sampling of the reference law itself.
Dividing by $s_{\mathrm{PF}}^2$ makes the error dimensionless.
Table~\ref{tab:ex3onestep} shows smaller diffusion-model errors in both amplitude groups.
The LSTM errors are close to those of the moment-matched Gaussian, indicating that the Gaussian output family limits conditional accuracy in this test.

\begin{table}[H]
\centering
\caption{Median one-step excess error for the SOL example.
Quiet and active histories have current signal amplitudes below the median and above the $80$th percentile, respectively ($64$ histories per group).
The moment-matched Gaussian uses the conditional mean and variance estimated from the particle-filter predictive law.}
\label{tab:ex3onestep}
\begin{tabular}{lcc}
\toprule
model & quiet & active \\
\midrule
memory + diffusion & $0.0050$ & $0.0113$ \\
moment-matched Gaussian & $0.185$ & $0.130$ \\
Gaussian-head LSTM & $0.187$ & $0.142$ \\
\bottomrule
\end{tabular}
\end{table}

\begin{figure}[htbp]
\centering
\exsolfig{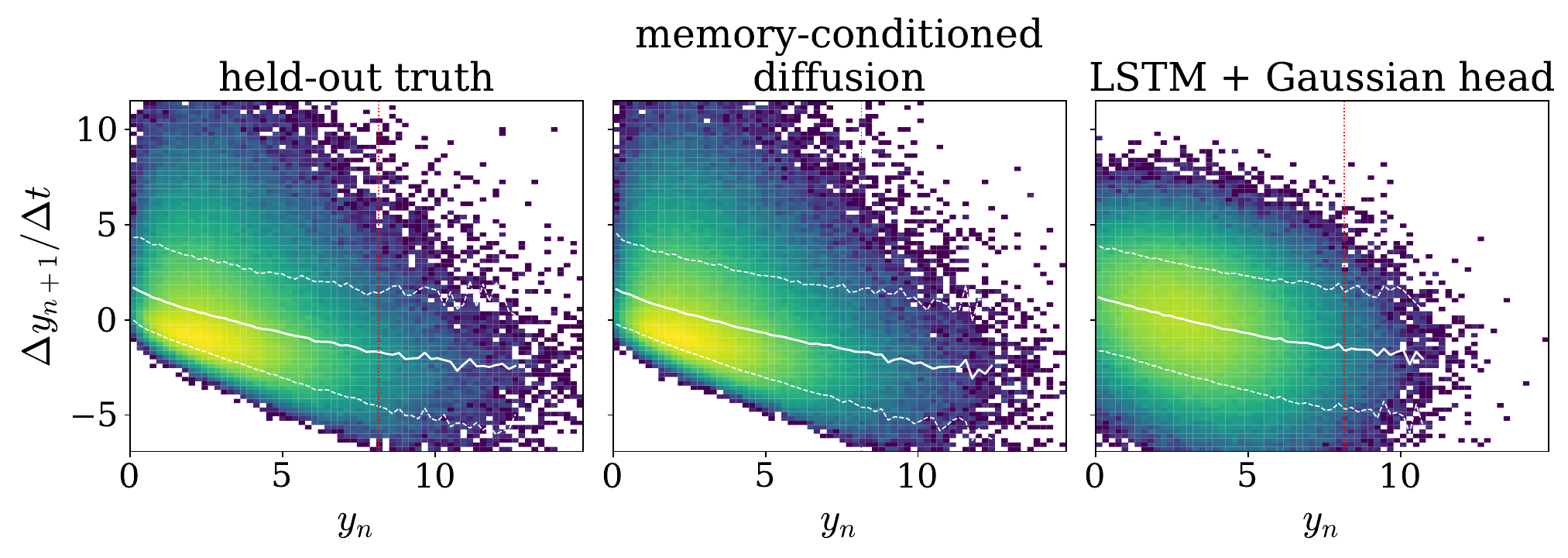}{0.9\textwidth}
\caption{Stochastic phase portrait of the held-out data and long model rollouts.
Color shows the joint density of $y_n$ and $\Delta y_{n+1}/\Delta t$ on a common logarithmic scale; white curves show the conditional mean and $10$--$90\%$ interval, and the red dotted line marks the burst threshold $\mu+2.5\sigma$.}
\label{fig:ex3phase}
\end{figure}

Figure~\ref{fig:ex3phase} compares the joint distributions of signal and increment in long rollouts.
New arrivals form the upper branch and decaying pulses the lower branch.
The diffusion model reproduces this asymmetry, while the LSTM narrows the upper branch and suppresses large positive increments.
A likely contributor is the LSTM's single Gaussian output head: at a fixed history, it represents the next increment by a symmetric distribution, whereas random pulse arrivals produce a skewed conditional law with a pronounced positive tail.
The similar one-step errors of the LSTM and the moment-matched Gaussian in Table~\ref{tab:ex3onestep} support this interpretation, although accumulated rollout errors can also affect the joint distribution.

\begin{figure}[htbp]
\centering
\exsolfig{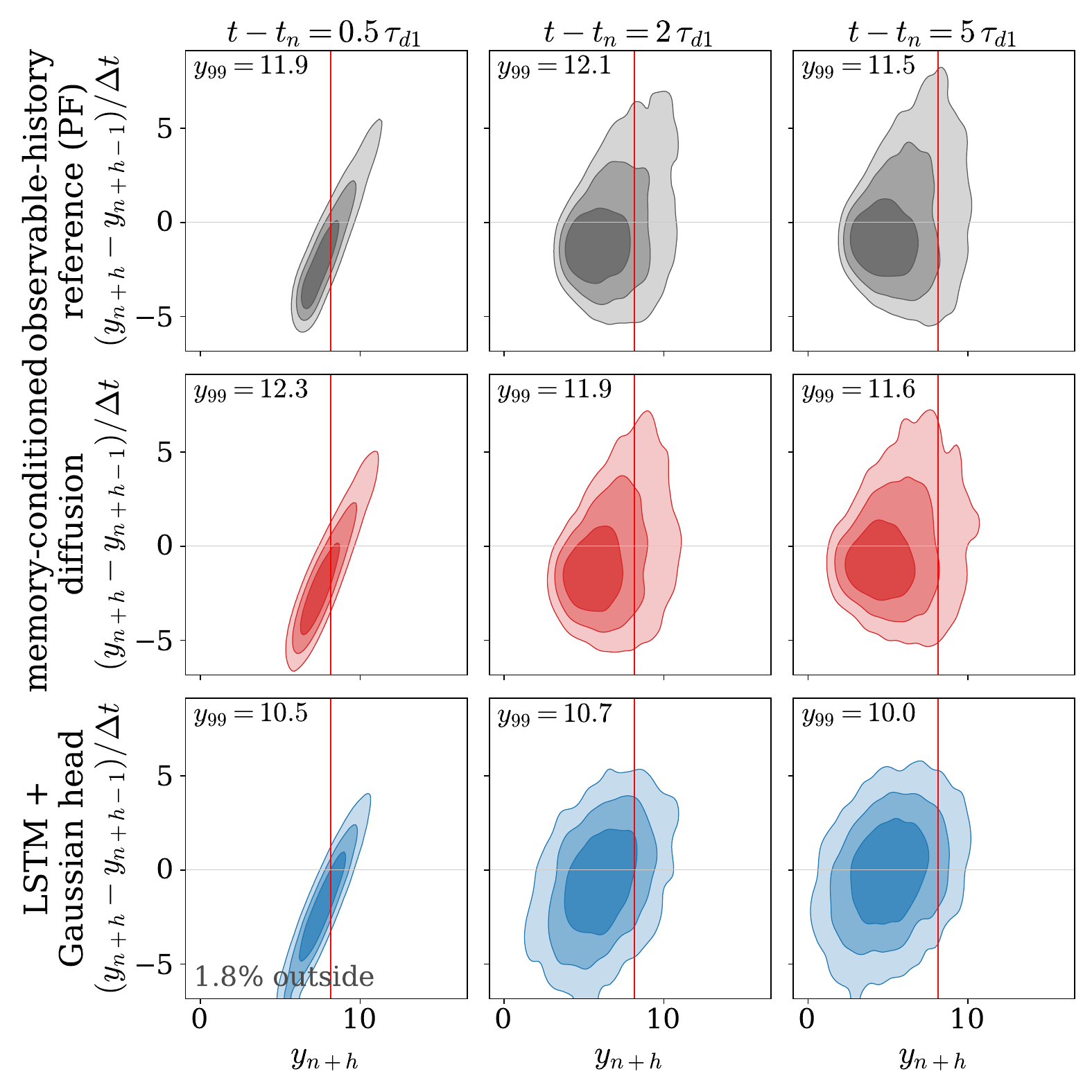}{0.76\textwidth}
\caption{Evolution of $4000$ paths released from the same active test history, shown at three forecast horizons in the phase plane.
Rows correspond to the particle-filter reference, the memory-conditioned diffusion model, and the Gaussian-head LSTM; contours enclose $50/80/95\%$ of the in-frame mass, the red line marks the burst threshold, and $y_{99}$ is the $99$th percentile of the amplitude.}
\label{fig:ex3cloud}
\end{figure}

Conditional forecasts from one active history show the same distinction (Figure~\ref{fig:ex3cloud}, $4000$ paths per model).
The diffusion model follows the reference ridge and upper tail during relaxation, while the LSTM produces a more elliptical cloud and underestimates the upper quantile.

\begin{figure}[htbp]
\centering
\exsolfig{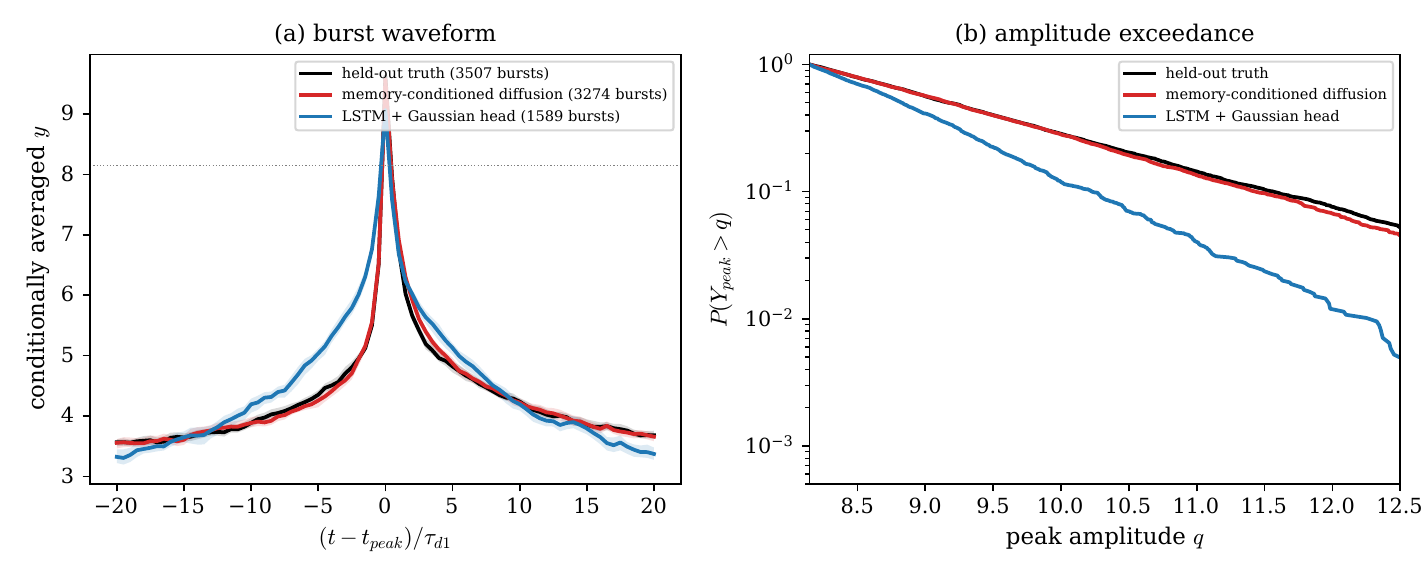}{0.85\textwidth}
\caption{Burst statistics of the held-out data and model rollouts.
(a) Conditionally averaged waveform of bursts above $\mu+2.5\sigma$, with $95\%$ trajectory-bootstrap bands. (b) Exceedance probability of the per-event peak amplitude; the extreme-event probability of Table~\ref{tab:ex3bursts} is instead a fraction of time samples and is not a peak statistic.}
\label{fig:ex3bursts}
\end{figure}

We next evaluate bursts, defined as contiguous excursions above $\mu+2.5\sigma$, where $\mu$ and $\sigma$ are the reference signal's stationary mean and standard deviation; chains of excursions whose consecutive peaks are separated by less than $2\tau_{d1}$ are merged into a single event, whose peak is the largest of the merged excursions.
Figure~\ref{fig:ex3bursts} shows that the diffusion model captures the rapid rise, slower decay, and broad peak-amplitude distribution.
Table~\ref{tab:ex3bursts} shows a burst rate close to the reference; the diffusion model's extreme-event probability has a lower point estimate than the reference, with marginally overlapping bootstrap intervals.
The LSTM produces fewer bursts, a broader peak-aligned average waveform, and a shorter peak-amplitude tail.

\begin{table}[H]
\centering
\caption{Burst statistics on $120$ records of duration $1500\,\tau_{d1}$ for each source, with $95\%$ trajectory-bootstrap intervals.
The extreme-event probability is the fraction of time samples with $y>\mu+5\sigma$.}
\label{tab:ex3bursts}
\setlength{\tabcolsep}{1pt}
\begin{tabular}{lccc}
\toprule
 & held-out data & memory + diffusion & Gaussian-head LSTM \\
\midrule
burst rate ($\times10^{-2}/\tau_{d1}$) & $1.99$ $[1.91, 2.06]$ & $1.88$ $[1.80, 1.97]$ & $0.91$ $[0.85, 0.97]$ \\
extreme-event prob.\ ($\times10^{-4}$) & $8.9$ $[7.1, 10.8]$ & $5.7$ $[4.5, 7.1]$ & $0.19$ $[0.05, 0.39]$ \\
\bottomrule
\end{tabular}
\end{table}

\begin{figure}[htbp]
\centering
\exsolfig{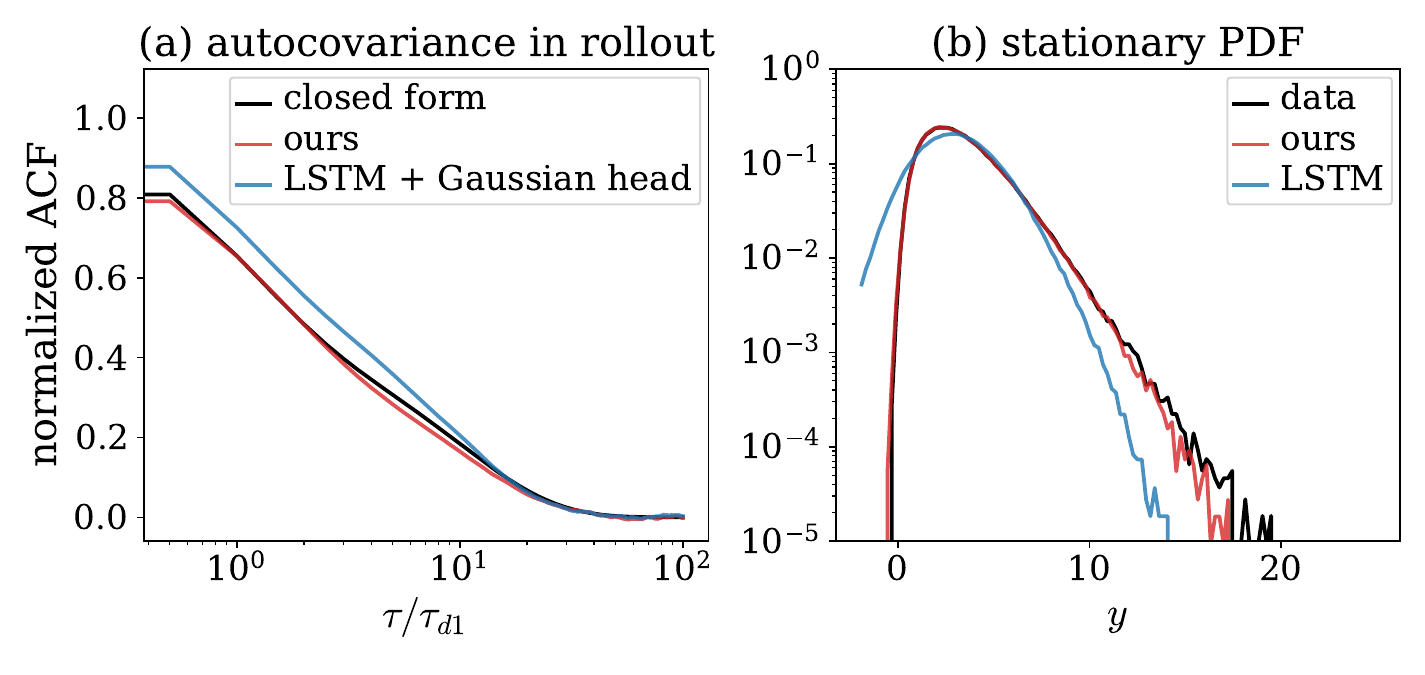}{0.8\textwidth}
\caption{Closed-loop statistics of the model rollouts; the displayed model curves use rollout seed~$0$ with $64$ trajectories of $8000$ steps.
(a) Normalized autocovariance compared with the closed form. (b) Stationary density compared with the held-out data.}
\label{fig:ex3roll}
\end{figure}

Long rollouts reproduce the skewed stationary density and both correlation scales with the diffusion model (Figure~\ref{fig:ex3roll}).
The LSTM captures the slow correlation tail but misses the distribution's upper tail.
These comparisons demonstrate the method's ability to reproduce non-Gaussian fluctuations as well as temporal dependence.

\section{Conclusions}
\label{sec:conclusions}

We developed a memory-conditioned diffusion method in which a compact, recursively updated filter bank enables generative flow maps to preserve long-time dependence without conditioning on long observation windows.
The bank summarizes resolved trajectories over multiple time scales, and predictive coordinates can reduce the dimension used by the conditional score estimator.
Conditional diffusion samples train a flow map whose evaluation and memory update have a cost independent of trajectory length for a fixed bank and network size.

The three examples demonstrate complementary capabilities.
In the scalar multiscale GLE, a compact filter bank retains the predictive information needed to reproduce slow correlations.
In the equilibrium vector GLE, compression to a two-dimensional increment predictor supports conditional sampling while the full bank maintains the recursive memory update, reproducing correlations induced by coupled, noncommuting memory modes.
This example uses reference-assisted bank selection, whereas Examples~1 and~3 use observation-based selection.
For the non-Gaussian SOL process, the method represents conditional asymmetry and intermittent-event statistics.
Together, these results illustrate the complementary roles of a compact history representation and an expressive conditional distribution.
The recurrent baselines provide context for these tests: the full-covariance Gaussian-head LSTM gives comparable correlation accuracy in the vector GLE, while the SOL comparison is limited by its Gaussian output family and does not establish a limitation of recurrent models with more expressive conditional distributions.

The present method assumes stationary, ergodic observations with fading memory that can be approximated by a modest filter bank.
The nonparametric sampler requires control of the conditioning dimension, while mean-based compression need not preserve non-Gaussian conditional laws.
Future work will consider richer memory representations and nonlinear predictive reductions for state-dependent and nonstationary dynamics.

\section*{Acknowledgments}

This material is based upon work supported in part by the U.S.~Department of Energy, Office of Science, Offices of Advanced Scientific Computing Research and Fusion Energy Science, and by the Laboratory Directed Research and Development program at the Oak Ridge National Laboratory, which is operated by UT-Battelle, LLC, for the U.S.~Department of Energy under Contract DE-AC05-00OR22725. S.\ He acknowledges support from the University of Tennessee, Knoxville AI seed grant.

\section*{Data availability}

The code for the numerical examples is available at \url{https://github.com/mlmathphy/DiffusionMemory_GLE2}.

\appendix
\section{Additional details on the memory representation}
\label{app:memory}

This appendix specifies memory selection, initialization, and predictive reduction, and relates the filters to exponential-memory variables.

\subsection{Filter conventions, initialization, and rate placement}
\label{app:selection}

We use the unnormalized convention in Eq.~\eqref{eq:ema}; the usual factor $1-\rho_i$ would only rescale component $i$ and is absorbed when the conditioning variables are whitened.
Unrolling the recursion,
\begin{equation}
\label{eq:emalinear}
m_n^{(i)} \;=\; \sum_{l=0}^{n-1} \rho_i^{\,l}\,
\bigl(X_{n-l} - X_{n-l-1}\bigr),
\end{equation}
the initialization transient decays as $\rho_i^n$.
For the selected bank, training begins at $n_{\mathrm{burn}}=\lceil-\log\delta/(\lambda_1\Delta t)\rceil$, with $\delta=0.03$, $0.001$, and $0.05$ in Examples~1--3, respectively; deployment initializes the bank from observed prehistory.

Within each candidate band $[\lambda_{\min},\lambda_{\max}]$ of Section~\ref{sec:features}, the rates are log-spaced: for $k\ge2$,
\begin{equation}
\label{eq:rates}
\lambda_i=\lambda_{\min}
\left(\frac{\lambda_{\max}}{\lambda_{\min}}\right)^{(i-1)/(k-1)},
\qquad i=1,\ldots,k,
\end{equation}
and for $k=1$ the geometric center $\lambda_1=\sqrt{\lambda_{\min}\lambda_{\max}}$.
These are history-feature rates, not estimates of physical decay rates.

Examples~1 and~3 search $k=1,\ldots,10$ and $k=1,\ldots,8$, respectively.
Both use eight logarithmically spaced lower endpoints over the interval in Section~\ref{sec:features} and upper endpoints $\{1,2,3\}/\Delta t$, retaining only bands with $\lambda_{\min}<\lambda_{\max}$.
The state-autocorrelation thresholds defining $\ell_{\mathrm{sig}}$ are $0.05$ and $0.02$, respectively.
Selection uses the first $20$ and $40$ training trajectories, split $60\%/40\%$ for fitting and validation.
Stage~A evaluates rates on a $40$-point logarithmic dictionary and retains bands within $2\%$ of the best validation risk at each $k$; Stage~B refits at the exact rates and requires $\max_l|C_{\mathrm{sur}}(l)-C_{\mathrm{data}}(l)|/C_{\mathrm{data}}(0)\leq0.05$, over lags $0$--$2000$ and $0$--$400$, respectively.
Here $C_{\mathrm{sur}}$ and $C_{\mathrm{data}}$ are the surrogate and selection-data autocovariances.
Example~2 uses the reference-based selection described in Section~\ref{sec:ex2gle}.

\subsection{Predictive rank and predictive coordinates}
\label{app:prank}
Correlated bank features can contain redundant predictive information.
We identify linear combinations that explain future variations beyond those predicted by the current state, and use their singular spectrum to define candidate reduced dimensions.

All quantities are linear residual covariances of centered variables, well defined for any second-order stationary process.
With $\bar m$, $\bar x$ the stationary means and $Q_x = \Sigma_{xx}^{\dagger}$ the Moore--Penrose inverse on the covariance support, put
\begin{equation}
\label{eq:resid}
\tilde m_n = (m_n - \bar m) - \Sigma_{mx} Q_x (x_n - \bar x),
\qquad
\tilde X_{n+h} = (X_{n+h} - \bar x) - \Sigma_{X_{n+h}x} Q_x (x_n - \bar x),
\end{equation}
for the residuals of the best linear predictors from the current state, and let $\Sigma_{m|x} = \mathrm{Cov}(\tilde m_n) = E_r \Lambda_r E_r^\top$ be its eigendecomposition on the effective support, $r = \operatorname{rank}(\Sigma_{m|x})$ and $\Lambda_r > 0$.
We take as whitening of the memory residual the canonical choice
\begin{equation}
\label{eq:whiten}
W_m \;:=\; \Lambda_r^{-1/2} E_r^\top \;\in\; \R^{r \times kd} ,
\qquad\text{so that}\quad
W_m \Sigma_{m|x} W_m^\top = I_r .
\end{equation}
Calling a matrix an \emph{admissible} whitening when it satisfies both $W_m \Sigma_{m|x} W_m^\top = I_r$ and $W_m = W_m E_r E_r^\top$ --- that is, when it does not act on $\ker \Sigma_{m|x}$ --- every admissible whitening is $O \Lambda_r^{-1/2} E_r^\top$ for an orthogonal $O$, and the choice among them leaves everything below invariant; for positive definite $\Sigma_{m|x} = L L^\top$ the Cholesky whitening $W_m = L^{-1}$ is admissible.
Equations~\eqref{eq:resid} and \eqref{eq:whiten} are population constructions; in finite samples $Q_x$ and $W_m$ are replaced by eigenvalue-floored, full-rank counterparts for numerical stability, so that the empirical predictive spectrum is a regularized estimator of the population one.
For a set of horizons $h = 1, \dots, H$ we collect the whitened residual cross-covariances of the memory with the future,
\begin{equation}
\label{eq:prank}
B \;=\; W_m
\bigl[\, g_1 \;\; g_2 \;\cdots\; g_H \,\bigr],
\qquad\text{with}\quad
g_h \;:=\; \mathrm{Cov}\bigl(\tilde m_n,\, \tilde X_{n+h}\bigr),
\end{equation}
estimated from the training trajectories by the same least-squares machinery as stage~A.
The singular values of $B = U \,\mathrm{diag}(\sigma_1, \sigma_2, \dots)\, V^\top$ form the \emph{predictive spectrum} of the bank: the square of each is the contribution of one orthogonal whitened memory direction to the linearly explained variance of the future, summed over horizons.
Retaining the $d_{\mathrm{pred}}$ leading left singular vectors defines the \emph{predictive memory coordinates}
\begin{equation}
\label{eq:zproj}
m_n^{\mathrm{pred}} \;=\; P\, \tilde m_n \in \R^{d_{\mathrm{pred}}},
\qquad\text{with}\quad
P \;:=\; U_{d_{\mathrm{pred}}}^\top\, W_m .
\end{equation}
The linearly explained variance discarded by this projection is measured by
\begin{equation}
\label{eq:epsp}
\epsilon_{d_{\mathrm{pred}}} \;=\; \max_{1 \le h \le H,\; \|b_h\| > 0}\;
\frac{\|b_h\|^2 - \|U_{d_{\mathrm{pred}}}^\top b_h\|^2}{\|b_h\|^2},
\qquad\text{with}\quad b_h := W_m\, g_h ,
\end{equation}
the worst-horizon fraction of memory-explained variance lost by the projection, the norms being Frobenius norms when the observable is multivariate.
The horizon count $H$ is set so that $H \Delta t$ covers the predictive-memory decay range; it need not extend to the full evaluation horizon of the closed-loop statistics.
The predictive spectrum supplies \emph{candidate} dimensions --- those at which it drops sharply --- so that no elbow heuristic is treated as a rank estimator on its own.

The coordinates serve two distinct purposes.
As a diagnostic, the predictive rank counts the linearly predictive directions resolved in the chosen bank; Example~1 reports it against a system whose hidden variables are known.
As a compression, they suggest replacing the full-bank conditioning $(x_n, m_n) \in \R^{(k+1)d}$ by
\begin{equation}
\label{eq:pipelinered}
\mathcal{H}_n \;\longrightarrow\; m_n \;\longrightarrow\; m_n^{\mathrm{pred}} = P \tilde m_n
\;\longrightarrow\; p\bigl(X_{n+1} \mid X_n, m_n^{\mathrm{pred}}\bigr) ,
\end{equation}
at conditioning dimension $d + d_{\mathrm{pred}}$, the full bank still being updated by Eq.~\eqref{eq:ema} and projected on read-out since $m_n^{\mathrm{pred}}$ is not autonomous.
The bank-update cost then scales with $kd$ while the dimension presented to the conditional model is $d + d_{\mathrm{pred}}$, so increasing $k$ need not force a proportional increase in the latter.
The predictive loss of Eq.~\eqref{eq:pipelinered} is evaluated in Example~1, whose deployed simulator retains the full bank; a mean-coordinate special case that includes the current state in the regression is deployed in Example~2 (Section~\ref{sec:ex2gle}); and the data-selected bank of Example~3 reduces to a single predictive coordinate, computed with the future one-step increments \emph{and their squares} over horizons $h=1,\dots,6$ as prediction targets, each residualized on the current signal and standardized to unit variance, so the retained direction is sensitive to both the conditional mean and the conditional spread.

\subsection{Conditional-mean sufficiency of a selected bank}
\label{app:suff}
Equation~\eqref{eq:memapprox} is a modeling approximation whose accuracy must be assessed.
Stability of the filter recursion alone does not imply stability of the coupled simulator, because generated increments depend on the current memory.
To assess the history information retained for conditional-mean prediction, we use the prediction risk of a conditioning set $\mathcal{C}$,
\begin{equation}
\label{eq:risk}
V(\mathcal{C}) \;=\;
\E\Bigl[\,\bigl\| X_{n+1} - X_n -
\E[\, X_{n+1} - X_n \mid \mathcal{C}\,] \bigr\|^2 \Bigr],
\end{equation}
writing $V_{\mathrm{markov}} = V(x_n)$, $V_{\mathrm{mem}}(k) = V\bigl((x_n, m_n)\bigr)$, and $V_{\mathrm{full}}$ for the full-history limit; stage~A of Section~\ref{sec:features} minimizes a held-out empirical estimate of the risk of the best \emph{linear} predictor given $(x_n, m_n)$; the population value of this linear-prediction risk upper-bounds $V\bigl((x_n, m_n)\bigr)$ and coincides with it only when the conditional mean is linear in the conditioning variables, while the finite held-out estimate is a proxy that need not preserve the bound.
When these risks are computable --- as in the linear-Gaussian benchmark of Example~1 --- and $V_{\mathrm{markov}} > V_{\mathrm{full}}$, we define the conditional-mean sufficiency diagnostic
\begin{equation}
\label{eq:epsk}
\epsilon_k \;=\;
\frac{V_{\mathrm{mem}}(k) - V_{\mathrm{full}}}
     {V_{\mathrm{markov}} - V_{\mathrm{full}}} \;\in\; [0, 1],
\end{equation}
with $\epsilon_k = 1$ indicating no improvement in optimal conditional-mean prediction over state-only conditioning and $\epsilon_k = 0$ indicating that the bank captures all history information relevant to that conditional mean.
In general, this diagnostic does not establish sufficiency for the full conditional law: history-dependent variance or higher-order distributional features may remain unresolved.
This is a benchmark diagnostic rather than a requirement of the method: $V_{\mathrm{full}}$ is unavailable in general, and the construction of Section~\ref{sec:features} never uses it.
A monotonicity property is worth recording alongside it: for nested banks, enlarging the conditioning state cannot increase the best achievable prediction risk under a fixed proper loss, although finite-data estimation becomes harder as the dimension grows.
This concerns the ideal predictor only and does not assert convergence of the complete learned simulator.

\subsection{Exact algebraic relation to exponential-memory embeddings}
\label{app:algebra}
This correspondence concerns one member of the problem class of Section~\ref{sec:problem}: an observed process governed by a generalized Langevin equation,
\begin{equation}
\label{eq:gle-embedding}
\dot X_t \;=\; a(X_t) \;-\; \int_{-\infty}^{t} K(t-s)\, X_s \dd s \;+\; F_t ,
\end{equation}
with drift $a : \R^d \to \R^d$, \emph{memory kernel} $K : [0,\infty) \to \R^{d \times d}$, and \emph{random force} $F_t$, a stationary zero-mean Gaussian process with covariance $R(t-s) = \E[F_t F_s^\top]$ \citep{ottobre2011asymptotic, mckinley2018anomalous}.
For a kernel with poles $\mu_j$, $K(t) = \sum_j a_j e^{-\mu_j t}$, the memory integral is carried by auxiliary variables obeying $\dot \zeta_j = -\mu_j \zeta_j + X_t$ \citep{ceriotti2010colored, baczewski2013numerical}, which are exponentially weighted averages of the past \emph{state}, whereas the features of Eq.~\eqref{eq:ema} are exponentially weighted averages of the past \emph{increments}; summation by parts converts one into the other.
The identity below does not require the equilibrium relation $R(t)=k_BT K(|t|)$ \citep{kubo1966fluctuation}; it applies to both thermal and independently prescribed forcing.

Fix a rate $\lambda_i$ of the bank, $\rho_i = e^{-\lambda_i \Delta t}$, and define the discrete auxiliary variable
\begin{equation}
\label{eq:zetadisc}
\zeta_{i,n}^{\Delta} \;:=\; \rho_i^{\,n}\, \zeta_{i,0}^{\Delta}
\;+\; \Delta t \sum_{l=0}^{n-1} \rho_i^{\,l}\, X_{n-1-l},
\qquad\text{equivalently}\qquad
\zeta_{i,n+1}^{\Delta} \;=\; \rho_i\, \zeta_{i,n}^{\Delta} + \Delta t\, X_n ,
\end{equation}
whose initial value carries the prehistory: $\zeta_{i,0}^{\Delta} = 0$ represents a truncated prehistory, while the stationary choice
\begin{equation}
\label{eq:zetainit}
\zeta_{i,0}^{\Delta} \;=\; \Delta t \sum_{l=0}^{\infty} \rho_i^{\,l}\, X_{-1-l}
\end{equation}
represents the history before $t = 0$, and makes $(\zeta_{i,0}^{\Delta}, X_0, X_{-1}, \dots)$ jointly stationary.
Equation~\eqref{eq:zetadisc} is the first-order (rectangle-rule) discretization of the exponential convolution $\zeta(t) = \int_{-\infty}^{t} e^{-\lambda_i (t-s)} X_s \dd s$, itself the auxiliary variable $\zeta_j$ of the Markovian embedding when $\lambda_i$ coincides with a kernel pole $\mu_j$.
Summation by parts applied to Eq.~\eqref{eq:emalinear} gives the identity
\begin{equation}
\label{eq:sbp}
\sum_{l=0}^{n-1} \rho_i^{\,l}\, X_{n-1-l}
\;=\;
\frac{X_n - m_n^{(i)} - \rho_i^{\,n} X_0}{1 - \rho_i} ,
\end{equation}
so that, for $\zeta_{i,0}^{\Delta} = 0$,
\begin{equation}
\label{eq:zetaexact}
\zeta_{i,n}^{\Delta}
\;=\;
\frac{\Delta t}{1 - \rho_i}\,
\bigl( X_n - m_n^{(i)} \bigr)
\;-\;
\frac{\Delta t\, \rho_i^{\,n}}{1 - \rho_i}\, X_0 .
\end{equation}
The relation between $(X_n, m_n^{(i)})$ and $\zeta_{i,n}^{\Delta}$ is therefore \emph{exact}, not approximate: given $X_0$, the pair determines the discrete auxiliary variable, the residual dependence being the explicitly displayed term $\propto \rho_i^{\,n}$, which is the initialization transient controlled by the burn-in.
Three distinct errors separate $\zeta_{i,n}^{\Delta}$ from the stationary continuous-time auxiliary variable $\zeta_j(t_n)$: the quadrature error of Eq.~\eqref{eq:zetadisc}; the truncation of the prehistory, i.e.\ the discrepancy between $\zeta_{i,0}^{\Delta} = 0$ and a stationary initial value, which decays as $\rho_i^{\,n}$; and the rate mismatch when $\lambda_i \neq \mu_j$.

The second memory channel is the correlated forcing.
The future realization of the force is not determined by the observed path; what the path determines is its conditional law, and this predictive information is what a memory summary must capture.
For linear-Gaussian models it is carried by a linear filter of the observed history with lag-decaying weights; such filters can be approximated by elements of the span of Eq.~\eqref{eq:emalinear}, and the accuracy of this approximation, rather than the kernel alone, governs the error of the approximation hypothesis Eq.~\eqref{eq:memapprox}.

\section{Reverse-ODE discretization and endpoint regularization}
\label{app:sampler}

The coefficients $b$ and $\sigma^2$ of Eq.~\eqref{eq:reverseode} diverge as $\tau \uparrow 1$, so the implementation evaluates them with the regularized schedule
\begin{equation}
\label{eq:regsched}
\alpha_\tau^{(h)} = 1 - \tau + h,
\qquad
(\beta_\tau^{(h)})^2 = \tau + h,
\qquad h = 1/n_{\mathrm{ODE}} .
\end{equation}
Writing $F_h(Z, \tau \mid c)$ for the right-hand side of Eq.~\eqref{eq:reverseode} evaluated with these coefficients and $\widehat S^{\mathrm{MC}}$, and $\tau_j = 1 - j \Delta\tau$ with $\Delta\tau = 1/n_{\mathrm{ODE}}$ for the backward grid, the state is advanced from a standard-normal terminal condition by
\begin{equation}
\label{eq:euler}
Z_{j+1} \;=\; Z_j - \Delta\tau\, F_h\bigl(Z_j, \tau_{j+1} \mid c\bigr),
\qquad j = 0, \dots, n_{\mathrm{ODE}} - 1 .
\end{equation}
This update is based on the explicit Euler scheme of \citep{liu2025trainingfree}, but evaluates the time-dependent coefficients at the lower node $\tau_{j+1}$ rather than at the state node $\tau_j$, which avoids $\tau_0 = 1$, where the regularized coefficients are stiffest ($|b| = 1/h$) and the unregularized ones are singular.
The regularized $\alpha_\tau^{(h)}$ and $\beta_\tau^{(h)}$ enter both the reverse drift and the score estimator; the score identity Eq.~\eqref{eq:exactscore} is the unregularized one.

The regularized schedule Eq.~\eqref{eq:regsched} is associated with the Gaussian corruption family
\begin{equation}
\label{eq:regtrans}
q^{(h)}_\tau(z \mid z_0) \;:=\;
\mathcal{N}\bigl(z;\, (1 - \tau + h)\, z_0,\; (\tau + h)\, I\bigr),
\end{equation}
which at $\tau = 1$ is $\mathcal{N}\bigl(h z_0, (1 + h) I\bigr)$, whereas Eq.~\eqref{eq:euler} is initialized from $\mathcal{N}(0, I)$, and which at $\tau = 0$ is the law of the blurred variable $(1 + h) Z_0^c + \sqrt{h}\, \Xi$ with $\Xi$ standard normal.
Because $q^{(h)}_0$ is already blurred, Eq.~\eqref{eq:regtrans} is not the transition law of a process started at $Z_0^c$; the regularization and the shifted-node discretization are coupled numerical approximations, and they do not exactly realize the reverse flow of this family.
Reading the endpoint blur in the original units, $\Delta X^{(h)} = (1 + h)\, \Delta X + \sqrt{h}\, \Xi / \kappa_s$, so that
\begin{equation}
\label{eq:blurcov}
\mathrm{Cov}\bigl(\Delta X^{(h)} \mid c\bigr)
\;=\;
(1 + h)^2\, \Sigma(c) \;+\; \frac{h}{\kappa_s^2}\, I ,
\qquad \Sigma(c) := \mathrm{Cov}\bigl(\Delta X \mid c\bigr) .
\end{equation}
Along a unit direction $v$, writing $V_v = v^\top \Sigma(c)\, v$, the relative variance increase is $2h + h^2 + h/(\kappa_s^2 V_v)$, whose final term is the additive endpoint blur and grows in relative importance along directions in which the conditional law is narrow.
In the error analysis of \citep{wang2026error} this inflation is an explicit parameter, separate from the integration step; tying the two together through $h = 1/n_{\mathrm{ODE}}$ means that refining the integration also shrinks the endpoint regularization, so the step-doubling check reported in Example~1 measures the combined change from both effects rather than either alone.

\section{Conditional score construction}
\label{app:score}
For $0<\tau<1$, Eq.~\eqref{eq:score} is the score of a Gaussian mixture whose component weights are determined by the conditioning kernel.
This construction is related to kernel conditional density estimation \citep{hall2004crossvalidation}.
Its accuracy depends on the conditioning dimension, bandwidth, and available trajectory data.

The Gaussian corruption in Eq.~\eqref{eq:forward} is generated by the auxiliary diffusion
\begin{equation}
\label{eq:forwardsde}
\dd Z_\tau^c=b(\tau)Z_\tau^c\dd\tau+\sigma(\tau)\dd B_\tau,
\end{equation}
where $B_\tau$ is a Brownian motion in diffusion time and $b,\sigma$ are defined in Section~\ref{sec:tfdiff}.

Because the diffused conditional is the Gaussian mixture
\begin{equation*}
p_\tau(z \mid c) = \int \mathcal{N}(z;\, \alpha_\tau z_0, \beta_\tau^2 I)\, p(z_0 \mid c) \dd z_0 ,
\end{equation*}
the score has an exact weighted-average representation \citep{liu2025trainingfree},
\begin{equation}
\label{eq:exactscore}
S(z,\tau \mid c) =
\int \frac{\alpha_\tau z_0 - z}{\beta_\tau^2}\,
w_\tau(z_0 \mid z, c) \dd z_0 ,
\end{equation}
where
\begin{equation*}
w_\tau(z_0 \mid z, c) :=
\frac{\mathcal{N}(z;\, \alpha_\tau z_0,\, \beta_\tau^2 I)\,
p(z_0 \mid c)}
{\displaystyle \int \mathcal{N}(z;\, \alpha_\tau \bar z_0,\,
\beta_\tau^2 I)\, p(\bar z_0 \mid c) \dd \bar z_0} .
\end{equation*}
Three steps turn Eq.~\eqref{eq:exactscore} into a data-driven estimator.
First, the observation data are samples of the joint law of $(c, z)$ rather than of the conditional law at any fixed query $c^*$, so the conditional is written in the normalized form
\begin{equation}
\label{eq:condid}
p(z_0 \mid c^*) \;=\;
\frac{\displaystyle\int \delta(\tilde c - c^*)\, p(z_0, \tilde c)\, \dd \tilde c}
     {\displaystyle\int \delta(\tilde c - c^*)\, p(\tilde c)\, \dd \tilde c} ,
\end{equation}
in which the sampling density of the conditioning variable appears explicitly.
Second, the Dirac delta is mollified into a Gaussian kernel of bandwidth $\nu$ in the conditioning variable and both integrals are replaced by empirical sums over the data tuples, which produces the self-normalized conditioning weights of Eq.~\eqref{eq:score}, following the conditional smoothing construction of \citep{liu2025trainingfree}.
Here the tuples are temporally dependent, so the nominal neighborhood count does not directly measure the amount of independent information it carries; we therefore choose $\nu$ as an empirical trade-off between conditioning bias and neighborhood size.
Third, the sum is restricted to the $J$ nearest neighbors of $c^*$, a computational truncation whose adequacy is monitored through the neighborhood radius and the effective sample size of the conditioning weights.
Combining these steps yields Eq.~\eqref{eq:score}.
The weight effective sample size measures concentration of the empirical weights, not the number of statistically independent observations.

\bibliographystyle{elsarticle-num}
\bibliography{refs}

\end{document}